\pdfoutput=1

\documentclass[11pt]{article}

\usepackage[final]{acl}

\usepackage{times}
\usepackage{latexsym}

\usepackage[T1]{fontenc}

\usepackage[utf8]{inputenc}

\usepackage{microtype}

\usepackage{inconsolata}

\usepackage{graphicx}

\usepackage{booktabs}
\usepackage{bbm} 
\usepackage{amsmath}
\usepackage{algorithm}
\usepackage{algpseudocode}
\usepackage{amsfonts}
\usepackage{amsthm}
\usepackage{amssymb}
\usepackage{subcaption}
\usepackage{makecell}
\usepackage{xspace}

\theoremstyle{plain}
\newtheorem{prop}{Proposition}

\theoremstyle{definition}
\newtheorem{defn}{Definition}

\theoremstyle{remark}

\DeclareMathOperator*{\argmax}{argmax}
\DeclareMathOperator*{\argmin}{argmin}

\usepackage[textsize=small,disable]{todonotes}

\newcommand{\bino}[0]{Binoculars\xspace{}}
\newcommand{\fast}[0]{FastDetectGPT\xspace{}}
\newcommand{\llama}[0]{Llama\xspace{}}
\newcommand{\falcon}[0]{Falcon\xspace{}}
\newcommand{\tower}[0]{Tower\xspace{}}
\newcommand{\towerbase}[0]{TowerBase\xspace{}}

\newcommand{\mosaic}[0]{MOSAIC\xspace}

\usepackage{xcolor}

\usepackage{array}
\newcolumntype{C}[1]{>{\centering\arraybackslash}p{#1}} 

\title{MOSAIC: Multiple Observers Spotting AI Content}

\author{
  \textbf{Matthieu Dubois\textsuperscript{1}} \and
  \textbf{François Yvon\textsuperscript{1}} \and
  \textbf{Pablo Piantanida\textsuperscript{2,3,4}}
\\
  \textsuperscript{1}Sorbonne Université, CNRS, ISIR, Paris France \\
  \textsuperscript{2} CNRS, International Laboratory on Learning Systems, Montréal, Canada \\
  \textsuperscript{3}Mila - Québec AI Institute, Montréal, Canada \\
  \textsuperscript{4}CentraleSupélec, Université Paris-Saclay, Gif-sur-Yvette, France
\\
  \small{
    \textbf{Correspondence:} \href{mailto:duboism@isir.upmc.fr}{duboism\;\textit{at}\;isir.upmc.fr}
  }
}

\begin{document}
\maketitle
\begin{abstract}
The dissemination of Large Language Models (LLMs), trained at scale, and endowed with powerful text-generating abilities, has made it easier for all to produce harmful, toxic, faked or forged content. In response, various proposals have been made to automatically discriminate artificially generated from human-written texts, typically framing the problem as a binary classification problem. Early approaches evaluate an input document with a well-chosen detector LLM, assuming that low-perplexity scores reliably signal machine-made content. More recent systems instead consider two LLMs and compare their probability distributions over the document to further discriminate when perplexity alone cannot. However, using a fixed pair of models can induce brittleness in performance. We extend these approaches to the ensembling of several LLMs and derive a new, theoretically grounded approach to combine their respective strengths. Our experiments, conducted with various generator LLMs, indicate that this approach effectively leverages the strengths of each model, resulting in robust detection performance across multiple domains. Our code and data are available at \url{https://github.com/BaggerOfWords/MOSAIC}.

\end{abstract}

\section{Introduction \label{sec:introduction}} 
Large Language Models (LLMs) 
have greatly improved the fluency and diversity of machine-generated texts. The release of ChatGPT and GPT4 by OpenAI has sparked global discussions regarding the new opportunities offered by AI-based writing assistants. These advances have also introduced considerable threats such as fake news generation \cite{zellers-etal-2019-DefendingNeuralFake}, and the potential for harmful outputs such as toxic or dishonest content \citep{crothers-etal-2023-machine}, among others. As it seems, the research on methods to detect the origin of a given text to mitigate the dissemination of forged content and to prevent technology-aided plagiarism still lags behind the rapid advancement of AI itself.\footnote{As illustrated by the discontinuation of OpenAI's detector \url{https://openai.com/index/new-ai-classifier-for-indicating-ai-written-text/}.} 

Many studies have focused on tools that could spot such AI-generated outputs and mitigate these underlying risks.
From a bird's eye view, this typically involves using \emph{detector} models to discriminate \emph{generator} models' outputs from legitimate human writings. Multiple versions of this generic text classification task have been considered, varying, e.g. the number of possible categories to distinguish and the amount of supervision (see Section~\ref{sec:related}). Owing to its large user base and applications, the largest effort has focused on one specific generator, ChatGPT, for which training and test data are easily obtained. Yet, the corresponding supervised binary problem, with a unique known generator, is not the only way to frame this task. A more challenging problem, that we study here, \textbf{is generator-agnostic artificial text detection}, where the models to be detected are not predefined.

\begin{figure}[t]
    \centering
    \includegraphics[width=\linewidth]{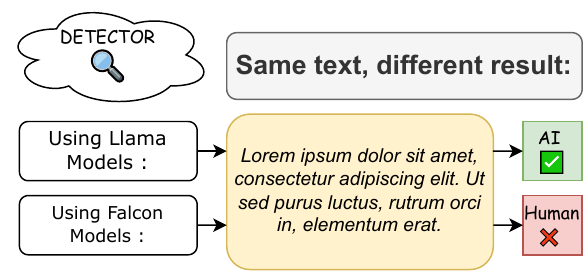}
    \caption{Unsupervised methods rely on fixed \textit{detector} models, but which one should you use in a \textbf{generator-agnostic} setting ?}
\end{figure}

\textbf{Our contributions.} In this paper, using fundamental information-theoretic principles from universal compression, we derive a new ensemble method (depicted in Figure~\ref{fig:architecture}) that combines the strength of multiple LLMs into a single score to detect forged texts. By using several models, this approach dispenses with the need to look for the optimal set of \textit{detector} models, and thus does not require a validation dataset. Our experiments use both standard benchmarks comprising multiple domains and languages, as well as a new corpus of machine-generated texts. They confirm that ensembling strong LLMs yields detectors that can robustly identify a multiplicity of generators and that compare favourably with several recent proposals using a predefined set of detector models. We also report a number a complementary analyses exploring the effect of incrementally growing the ensemble, of considering adapted versions of one single model, of using a naive ensembling method, and of processing noised versions of artificial texts. We publicly release the code and data produced for this study.

\section{Related Work\label{sec:related}}

\textbf{Overview of the field}. The improved text generation abilities of LLMs raise concerns about potential misuses such as disinformation \citep{zellers-etal-2019-DefendingNeuralFake}, abusive content \citep{crothers-etal-2023-machine}, forged academic publications \citep{liu-etal-2024-onthedetectability}, or cheating during exams \citep{vasilatos-etal-2023-HowkGPTInvestigatingDetection}. Since such fake texts seem difficult for humans to spot \citep{gehrmann-etal-2019-gltr}, the issue of automatically detecting machine-generated texts has been subject to an increasing focus. This problem can be framed as a binary human vs.\ non-human decision, as the problem of detecting one known artificial agent (e.g., ChatGPT \citep{mitrovic-etal-2023-ChatGPTHumanDetect,liu-etal-2024-onthedetectability}), or as discriminating the correct model in a predefined list \citep{liOriginTracingDetecting2023}. Some works aim to differentiate between ``machine-generated'' or ``machine-polished'' \cite{abassy-etal-2024-llm,liu-etal-2024-onthedetectability}. Another distinction is between closed-domain (e.g.\ scientific \citep{liyanage-etal-2022-benchmark}, academic \citep{liu-etal-2024-onthedetectability} or user-generated content \citep{fagni-etal-2021-TweepFakeDetectingDeepfake,kumarageStylometricDetectionAIGenerated2023}) vs.\ open-domain text detection. Assuming the generator models are known, various settings can be considered, depending on whether models can be openly queried (open parameters), whether they expose their full logits, or just the top prediction (and associated probability), etc.

\textbf{Supervised methods}. Supervised detection with a single generator often achieves detection accuracy rates in the high 90s \citep{zellers-etal-2019-DefendingNeuralFake,guo-etal-2023-HowCloseChatGPT,liu-etal-2024-onthedetectability}, using classifiers based on Roberta \citep{conneau-etal-2020-unsupervised} or T5 \citep{raffelExploringLimitsTransfer2020}. 
However, these approaches are brittle and their success depends on particular generator-detector pairs \citep{antoun-etal-2024-text}, 
prompting e.g.\ \citet{verma-etal-2024-ghostbuster} to design automatic feature extractors from multiple detectors to improve the robustness of their system.

\textbf{Zero-shot methods}. Unsupervised detection is more challenging. Most approaches rest on the idea that human-written texts are more ``surprising'' than artificial texts\footnote{Assuming generation does not use random sampling, in which case the reverse is likely to be observed, as long artificial texts drift away from natural writings \citep{zellers-etal-2019-DefendingNeuralFake}.}, leading to a difference in token-wise log-probability\footnote{\citep{mitchell-etal-2023-DetectGPT} argues that the difference is better seen at the level of log-ranks.}. This idea is used in GPTzero\footnote{\url{https://gptzero.me/}} and thresholding perplexity usually provides strong baselines (see, \textsl{inter alia}, \citep{gehrmann-etal-2019-gltr, ippolito-etal-2020-automatic, mitchell-etal-2023-DetectGPT}). Such techniques heavily rely on the \emph{detector} model(s) used to compute the log-probabilities of input texts, which must be robust to variations in domains, genres, styles, and languages \citep{wang-etal-2024-m4}; and to variations in the generator itself \citep{antoun-etal-2024-text}.

\textbf{Perturbation-based}. \citet{mitchell-etal-2023-DetectGPT} and \citet{bao-etal-2024-fastdetectgpt} exploit a similar intuition,
arguing that small random perturbations of an artificial text will on average make it less likely, unlike human-written texts. They develop a statistical criterion based on the curvature of the log-probability function and achieve near-perfect detection scores across three types of texts generated by five different models.
The \bino{} score of \citet{hans-etal-2024-SpottingLLMsBinoculars} also relies on a function of the per-token log-perplexity, contrasted with the cross-entropy of an auxiliary model.

These valuable works point to \textbf{the over-reliance on one specific detector model as a major limitation of the state-of-the-art.} Our proposed mitigation relies on ensemble techniques, that are also considered in the supervised detection setting, e.g.\ in \citep{verma-etal-2024-ghostbuster,wang-etal-2023-seqxgpt,el-sayed-nasr-2023-ensemble,liyanage-buscaldi-2023-ensemble}.

\textbf{Other methods}. Abandoning generator-detector based techniques altogether, \citep{mao-etal-2024-raidar,yang-etal-2024-dnagpt} develop effective detection approaches based on regeneration, prompting the (known) generator to regenerate part of the input text. The intuition is that artificial inputs are likely to be regenerated exactly, unlike human texts.
Other strategies include text watermarking \citep{kirchenbauer-etal-2023-awatermark,kirchenbauer-etal-2023-onthereliability,liu-etal-2024-AdaptiveTextWatermark}, though its efficiency and robustness are still subject to discussions, e.g., \citep{zhang-etal-2023-WatermarksSandImpossibility}.

\textbf{Robustness issues}. Recent works focus on detection robustness. \cite{wang-etal-2024-stumbling} find that after simple modifications, only watermarking remains able to accurately identify artificial documents. \citet{dugan-etal-2024-raid} present artificial texts generated with multiple models and sampling strategies, additionally subject to various adversarial attacks, observing that most detectors suffer large drops in performance, and \bino{} \citep{hans-etal-2024-SpottingLLMsBinoculars} stands out, achieving decent True Positive Rates at False Positive Rates under 1\%.

\begin{algorithm*}[t]
\caption{\mosaic{} Scoring} \label{alg:mosaic}
\begin{algorithmic}[1] 
\State \textbf{Input}:  text $\mathbf{y} =  \langle y_0, y_1, \ldots \rangle$, LLMs $m \in \mathcal{M}$, with $m^{\star}$ the reference model
\For{$y_t$ in $\mathbf{y}$}
    \State $\mu^\star(m|\mathbf{y}_{<t}) \gets \text{Blahut–Arimoto }(\mathcal{P}_{\mathcal{M}}(\mathcal{Y}); \mathbf{y}_{<t})$ \Comment{Obtain the $\mu^{\star}$ weights}
    \State $q^\star(y_t | \mathbf{y}_{<t}) \gets \sum\limits_{m \in \mathcal{M}} \mu^\star(m|\mathbf{y}_{<t})  p_{m} ( y_t |\mathbf{y}_{<t})$ \Comment{Build the mixture}
    \State $s_t(\mathbf{y}) \gets \mathcal{L}_{q^\star}(y_t | \mathbf{y}_{<t})  - {\sum\limits_{y\in\Omega}} \, p_{m^\star} ( {y} \mid \mathbf{y}_{<t}) \mathcal{L}_{q^\star}({y} \mid \mathbf{y}_{<t})$ \hspace{-2em}\Comment{Compare surprisal and cross-entropy}
\EndFor
\State $ S_{\mathcal{M}} (\mathbf{y}) \gets \frac{1}{T} \sum_{t} s_t(\mathbf{y})$ \Comment{\mosaic{} score for the whole text}
\end{algorithmic}
\end{algorithm*}

\section{Detecting AI-Generated Text with Multiple Models}
\subsection{Background}
We consider models for language generation tasks that define a probability distribution over strings. Formally, language models are probability distributions over an output space $\mathcal{Y}$ which contains 
all possible strings over vocabulary $\Omega$:
$ 
\mathcal{Y} \triangleq \big\{ \texttt{BOS}  \circ \mathbf{y}  \circ \texttt{EOS}\,  | \,  \mathbf{y} \in \Omega^* \big\},   
$ 
$\texttt{BOS}$ and $\texttt{EOS} $ denote respectively the beginning-of-sequence and end-of-sequence tokens, and $\Omega^* $ is the Kleene closure of $\Omega$.

Today’s models for language generation are typically parameterized with trainable weights $\theta\in\Theta$. These models follow a local-normalization scheme, meaning that $\forall$ $t > 0$, $ p_\theta (\cdot | \mathbf{y}_{<t}, )$ defines a conditional probability distribution over $\bar{\Omega} = \Omega \cup \texttt{EOS}$. The probability of a sequence $\mathbf{y}  = \langle y_0,  \ldots, y_T \rangle$ is:
\begin{equation}
    p (\mathbf{y}) = \prod_{t=1}^{T} 
   p_\theta ( y_t | \mathbf{y}_{<t})
    \label{eq:sequence_probability-eq}
\end{equation}
$\mathbf{y}_{<t} = \langle y_0, \ldots, y_{t-1} \rangle$, $y_0 = \texttt{BOS}$ and $y_T = \texttt{EOS}$. 

\textbf{Measuring information.} A fundamental relationship in information theory relates the probability of a message and the quantity of information it carries, using the relationship \citep{Cover91}: 
$
    \textrm{information} = -\log (\textrm{probability}), 
$
assuming the use of coding techniques such as Huffman and Arithmetic codes \citep{shields1996ergodic} which allow to achieve message lengths closely approximating this ideal length in binary digits.

\textbf{Explanations of data.} Given a body of text represented in a finite string $\mathbf{y}_{<t} = \langle y_0, \ldots, y_{t-1} \rangle$, an ``explanation'' of this next token $y_t$ is a binary string encoding the symbol with \emph{minimum} length $\mathcal{L}_p ( y_t | \mathbf{y}_{<t}) \triangleq  -\log p ( y_t | \mathbf{y}_{<t})$. $\mathcal{L}_p ( y_t | \mathbf{y}_{<t})$ is also often referred to as the model's \textbf{surprisal} \citep{samson-1953} on input $y_t$. Its expected value is termed the \textbf{conditional entropy:}
$$
\mathcal{H}_p (Y_t  | \mathbf{y}_{<t}) =  \sum_{y_t \in \Omega} p (y_t|\mathbf{y}_{<t}) \mathcal{L}_p (y_t | \mathbf{y}_{<t}). 
$$
Finally, another important concept is the \textbf{conditional mutual information} (MI) between two random variables $\mathbb{M}$ and $Y_t$, given a sequence value $\mathbf{y}_{<t}$, defined as \citep{Cover91}:
\begin{align*}
\mathcal{I}_p(\mathbb{M};Y_t| \mathbf{y}_{<t} ) &= \mathcal{H}_p (Y_t| \mathbf{y}_{<t} ) - \mathcal{H}_p(Y_t| \mathbb{M},\mathbf{y}_{<t} ),\\ 
\mathcal{H}_p (Y_t| \mathbb{M},\mathbf{y}_{<t} ) &= \underset{m \sim \mu (m | \mathbf{y}_{<t} ) }{\mathbb{E}}  \mathcal{H}_p (Y_t  | m,\mathbf{y}_{<t}) .
\end{align*}
Conditional MI captures the amount of information we get about $\mathbb{M}$ when observing $Y_t$, and already knowing $\mathbf{y}_{<t}$.

\subsection{Multi-model Detection Methods}

When detecting machine-generated texts in a zero-shot setting, the most promising methods rely on a language model \citep{guo-etal-2023-HowCloseChatGPT}. These techniques are becoming less effective as LLMs’ capabilities improve over time. The results of \fast{} and \bino{} suggest that detection can be significantly improved by simultaneously using two models: in their study, detection scores are obtained by comparing a model’s surprisal against the cross-entropy with respect to the other model, averaged over the input tokens. 

Here, we explore a key question: \textbf{how can we leverage multiple models for improved detection?} A straightforward approach is to systematically search for the best model pair, achieving optimal detection scores, as reported in Table~3 p.15 in \citep{hans-etal-2024-SpottingLLMsBinoculars} and Table~7 p.19 in \citep{bao-etal-2024-fastdetectgpt}. However, this method lacks robustness, as the best-performing model pair may vary depending on the validation dataset used for selection, leading to performance fluctuations when the test domain or language changes. Additionally, this approach may struggle to scale to larger model ensembles due to the exponential increase in possible model combinations that must be explored.

Before addressing our main question, we revisit and reformulate \bino{} and \fast{}\footnote{We focus on \bino{} here, while the analysis of \fast{} is provided in Appendix~\ref{ann:theory-similarities}.} using the previously introduced concepts, we explore potential variations and extensions.


\subsection{Revisiting the \bino{} Method \label{ssec:formulas}}
The \bino{} score $B_{p,q}(\mathbf{y})$ for an input sequence $\mathbf{y} =  \langle y_0, y_1, \ldots \rangle$, using two language models $q$ and $p$ expressed as  in \eqref{eq:sequence_probability-eq}, is defined by : 
\begin{equation}
    B_{p,q}(\mathbf{y}) \triangleq   \frac{ \sum_{t=1}^{T}  \sum_{y\in\Omega} \mathbbm{1} [y=y_t]
  \mathcal{L}_q(y_t |\mathbf{y}_{<t}) }{ 
   \sum_{t=1}^{T} 
    \sum_{y\in\Omega} \displaystyle  p (y | \mathbf{y}_{<t})\mathcal{L}_q(y | \mathbf{y}_{<t})}, 
    \label{eq:binoculars-score}
\end{equation}
with $\mathcal{L}_q ( y_t | \mathbf{y}_{<t}) = - \log q( y_t | \mathbf{y}_{<t})$, and $p (y | \mathbf{y}_{<t})$ and $q (y | \mathbf{y}_{<t})$ represent the probabilities assigned by models $p$ and $q$, respectively, to token $y$ conditioned on the current context $\mathbf{y}_{<t}$. It is important to note that Eq.~\eqref{eq:binoculars-score} is only valid when $q$ and $p$ share the same underlying vocabulary and tokenizer.

From an information theory perspective, we can interpret the numerator as the shorter average encoding length of the observed text according to model $q$, while the denominator represents the encoding length if the text were generated by sampling from model $p$ instead. For this reason, in all that follows, we call $p$ the \textbf{reference model}.

Interestingly, it is easy to see that \fast{} leverages the same concept but calculates a difference rather than a ratio. While equivalent, it normalizes the score for each token instead of averaging it over the entire sentence (see Appendix~\ref{ann:theory-similarities}).

\textbf{How to choose the most promising reference model?} Both scoring methods are based on the intuition that the numerator—the log-probability of the text—tends to be smaller for machine-generated texts compared to natural ones. To enhance these distinctions, they also rely on the idea that, conversely, the denominator term should be smaller for artificial texts.
This hints at the fact that when having a family of models $\mathcal{P}_{\mathcal{M}}(\mathcal{Y})=\{    p_m (\mathbf{y}) :\, m\in\mathcal{M}\}$, given a  human sample of text  $\mathbf{y}_{\textrm{hum}}$, we can use the following criterion:
\begin{equation}
 m^\star (\mathbf{y}_{\textrm{hum}})\triangleq  \argmin_{m\in \mathcal{M}} -\sum_{t=1}^{T} \log p_{m} ( y_t |\mathbf{y}_{<t}). 
 \label{eq-selection-m}
   \end{equation}
In other words, the reference model $p_{m^\star}$ needs to be the model in the ensemble $ \mathcal{P}_{\mathcal{M}}(\mathcal{Y})$ with the lowest log-perplexity for human samples of text $\mathbf{y}$. This often turns out to be the largest LLM in the ensemble, which is consistent with the tables in the original papers \citep{bao-etal-2024-fastdetectgpt,hans-etal-2024-SpottingLLMsBinoculars} and confirmed experimentally in Section~\ref{ssec:choosing-m}. 

The methodology introduced in \eqref{eq-selection-m} enables us to select the most promising reference model, $p_{m^\star}$, from a given family of available models $ \mathcal{P}_{\mathcal{M}}(\mathcal{Y})$. However, it does not provide a practical criterion for selecting the best model $q$ needed to evaluate \eqref{eq:binoculars-score}. This will be addressed in the next section.

\begin{table*}[htbp]
    \centering
    \small
    \setlength{\tabcolsep}{1pt}
    \begin{tabular}{l|C{1.24cm}|C{1.20cm}|C{1.20cm}|C{1.22cm}|C{1.24cm}|C{1.24cm}|C{1.24cm}|C{1.24cm}|C{1.22cm}|C{1.24cm}|C{1.21cm}}
    \toprule
     & \textbf{chatgpt} & \textbf{cohere-c} & \textbf{cohere} & \textbf{gpt2} & \textbf{gpt3} & \textbf{gpt4} & \textbf{llama-c} & \textbf{mistral-c} & \textbf{mistral} & \textbf{mpt-c} & \textbf{mpt} \\
    \midrule
    \textbf{Bino (best)} & 
    0.996  & 
    0.986  & 
    0.986  & 
    0.935  & 
    0.999  & 
    0.969  & 
    1.000  & 
    0.999  & 
    0.953  & 
    0.997  & 
    0.976 \\ 

    \textbf{Config.} & 
    7b/7b-i & 
    40b/7b-i & 
    40b/7b-i & 
    7b/7b-i & 
    40b/7b-i & 
    40b/40b-i & 
    7b/7b-i & 
    40b/7b-i & 
    40b/7b-i & 
    7b/7b-i & 
    40b/7b-i\\ \hline

    \textbf{Bino (min)} & 
    0.650 & 
    0.671 & 
    0.606 & 
    0.207 & 
    0.871 & 
    0.293 & 
    0.826 & 
    0.668 & 
    0.436 & 
    0.698 & 
    0.473 \\ \hline

    \textbf{Bino (avg)} & 
    0.893 & 
    0.894 & 
    0.874 & 
    0.616 & 
    0.971 & 
    0.675 & 
    0.967 & 
    0.925 & 
    0.755 & 
    0.935 & 
    0.795 \\ \hline\hline

    \textbf{Fast (best)} & 
    0.996 & 
    0.977 & 
    0.982 & 
    0.947 & 
    0.996 & 
    0.972 & 
    1.000 & 
    0.998 & 
    0.954 & 
    0.995 & 
    0.985 \\ 

    \textbf{Config.} & 
    40b/40b-i & 
    40b/7b-i & 
    40b/7b-i & 
    7b/7b-i & 
    40b/40b-i & 
    40b/40b-i & 
    40b/40b-i & 
    40b/7b-i & 
    40b/7b-i & 
    40b/40b-i & 
    40b/7b-i\\ \hline

    \textbf{Fast (min)} & 
    0.512 & 
    0.433 & 
    0.504 & 
    0.366 & 
    0.477 & 
    0.343 & 
    0.693 & 
    0.438 & 
    0.522 & 
    0.360 & 
    0.641 \\ \hline

    \textbf{Fast (avg)} & 
    0.848 & 
    0.816 & 
    0.834 & 
    0.680 & 
    0.859 & 
    0.681 & 
    0.932 & 
    0.850 & 
    0.781 & 
    0.826 & 
    0.853 \\
    \bottomrule
    \end{tabular}
    \caption{Summary of Bino(culars) \& Fast(DetectGPT) AUROC on RAID with the \falcon{} family. 
    Best, avg and min cells respectively report the best, average and worst score among all configurations.``-i'' and ``-c'' respectively denote the instruct and chat version of the model.}
    \label{tab:auc-falcon-min-max}
\end{table*}

\begin{table*}[htbp]
    \centering
    \small
    \setlength{\tabcolsep}{1pt}
    \begin{tabular}{l|C{1.24cm}|C{1.20cm}|C{1.20cm}|C{1.22cm}|C{1.24cm}|C{1.24cm}|C{1.24cm}|C{1.24cm}|C{1.22cm}|C{1.24cm}|C{1.21cm}}
    \toprule
     & \textbf{chatgpt} & \textbf{cohere-c} & \textbf{cohere} & \textbf{gpt2} & \textbf{gpt3} & \textbf{gpt4} & \textbf{llama-c} & \textbf{mistral-c} & \textbf{mistral} & \textbf{mpt-c} & \textbf{mpt} \\
    \midrule
    \textbf{Bino (best)} & 
    0.996  & 
    0.985  & 
    0.979  & 
    0.812  & 
    0.999  & 
    0.969  & 
    1.000  & 
    0.998  & 
    0.915  & 
    0.999  & 
    0.946 \\ 

    \textbf{Config.} & 
    T13b/L-c  & 
    L/L-c  & 
    T13b/L-c  & 
    T13b/T7b  & 
    T13b/L-c  & 
    T13b/L-c  & 
    L/L-c  & 
    T13b/L-c  & 
    T13b/T7b  & 
    T7b/L-c  & 
    T13b/T7b \\ \hline

    \textbf{Bino (min)} & 
    0.511 & 
    0.688 & 
    0.711 & 
    0.459 & 
    0.945 & 
    0.376 & 
    0.741 & 
    0.560 & 
    0.609 & 
    0.661 & 
    0.637 \\ \hline

    \textbf{Bino (avg)} & 
    0.837 & 
    0.900 & 
    0.870 & 
    0.652 & 
    0.983 & 
    0.720 & 
    0.928 & 
    0.876 & 
    0.774 & 
    0.895 & 
    0.798 \\ \hline\hline

    \textbf{Fast (best)} & 
    0.994 & 
    0.981 & 
    0.979 & 
    0.858 & 
    0.996 & 
    0.974 & 
    1.000 & 
    0.993 & 
    0.923 & 
    0.995 & 
    0.966 \\

    \textbf{Config.} & 
    T13b/L-c  & 
    L/L-c  & 
    L/L-c  & 
    T7b/L  & 
    L/L-c  & 
    T13b/L-c  & 
    L/L-c  & 
    T13b/L-c  & 
    T13b/L-c  & 
    T13b/T7b  & 
    L/L-c \\ \hline

    \textbf{Fast (min)} & 
    0.505 & 
    0.673 & 
    0.705 & 
    0.501 & 
    0.914 & 
    0.363 & 
    0.740 & 
    0.552 & 
    0.606 & 
    0.647 & 
    0.636 \\ \hline

    \textbf{Fast (avg)} & 
    0.802 & 
    0.853 & 
    0.860 & 
    0.684 & 
    0.961 & 
    0.691 & 
    0.918 & 
    0.869 & 
    0.796 & 
    0.866 & 
    0.830 \\
    \bottomrule
    \end{tabular}
    \caption{Summary of Bino(culars) \& Fast(DetectGPT) AUROC on RAID with \llama{} and \tower{} models. 
    Best, avg, and min cells report respectively the max, average and worst score among all configurations. ``T'' and ``L'' respectively stand for \towerbase{} and \llama-2-7b, while ``-c'' denotes the ``chat'' version.}
    \label{tab:auc-llama-min-max}
\end{table*}

\section{Introducing \mosaic{} \label{sec:methods}}

\begin{figure}[ht]
    \centering
    \includegraphics[width=\linewidth]{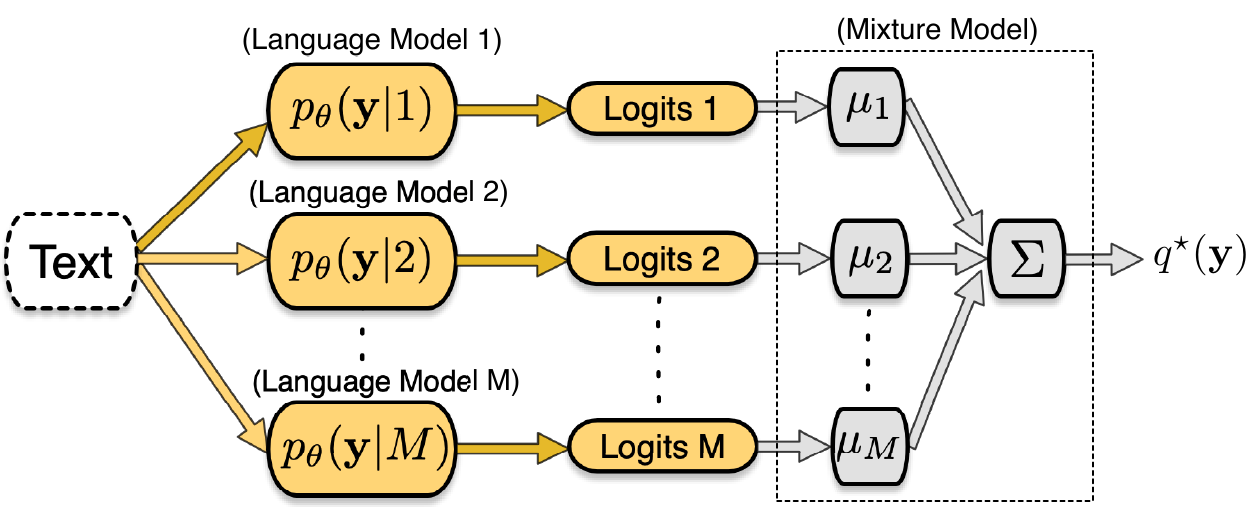}
    \caption{Mixture model, where $\{ \mu_i \}$ denote the time-varying weights associated to LLMs in the mixture.}
    \label{fig:architecture}
\end{figure}

Building on the principles used in previous systems, we now present \mosaic{}, a scoring method designed to leverage multiple models simultaneously. The key difference compared to previous approaches is that instead of using a single fixed LLM for $q$, \mosaic{} defines it as \textbf{a position-dependent mixture of all  models in the ensemble}. The weights of this mixture are formally defined in the next proposition  and depicted in Figure~\ref{fig:architecture}. 

\begin{prop}[Optimal model]\label{proposition}
The optimal model, $q^{\star}$—which minimizes the encoding length for tokens—is the position-dependent mixture:
\begin{equation}
q^\star( y_t | \mathbf{y}_{<t}) \triangleq \sum\limits_{m\in\mathcal{M}} \mu^\star(m|\mathbf{y}_{<t} ) p_m ( y_t | \mathbf{y}_{<t}),  
\end{equation}
where the distribution $\mu^\star(\cdot |\mathbf{y}_{<t})$ of the random variable $\mathbb{M}$ over LLM indices in $\mathcal{M}$ satisfies:  
\begin{equation}\label{eq-optimal-weights}
\mu^\star(\cdot |\mathbf{y}_{<t}) \triangleq  \argmax_{ \mu \in  \mathcal{P}(\Omega) }\,  \mathcal{I}_p \big(\mathbb{M} ; Y_t | \mathbf{y}_{<t} \big). 
\end{equation}
Furthermore, the weights $\{\mu^\star(m|\mathbf{y}_{<t})\}_{m\in\mathcal{M}}$ depend on the prefix $\mathbf{y}_{<t}$; they can be efficiently computed using the well-known Blahut–Arimoto algorithm \citep{Arimoto, Blahut}.
\end{prop}
\begin{defn}[\mosaic{} Score]\label{def:score}
 For an  input sentence $\mathbf{y} =  \langle y_0, y_1, \ldots \rangle$, and models indexed by $\mathcal{M}=\{ 1,\dots, M\}$ sharing a common tokenizer, the \mosaic{} score is then defined as: 
\begin{align}
 S_{m^\star,\mathcal{M}} (\mathbf{y}) \triangleq \frac{1}{T} \sum_{t=1}^{T} \sum_{y \in \Omega} 
\Bigg[  
    &\underbrace{\mathbbm{1}_{\{y = y_t\}} \mathcal{L}_{q^\star}(y_t \mid \mathbf{y}_{<t})}_{\textrm{(codelength for observed token)}}  \notag \\
    &\hspace{-10em} - \hspace{-1em}
    \underbrace{p_{m^\star} ( y_t \mid \mathbf{y}_{<t}) \mathcal{L}_{q^\star}(y_t \mid \mathbf{y}_{<t})}_{\textrm{(codelength for generated token from model } m)}
\Bigg] \label{eq:mosaic_score}
\end{align}
where $\mathcal{L}_{q^\star} ( y_t | \mathbf{y}_{<t}) = - \log q^{\star}( y_t | \mathbf{y}_{<t})$ and $m^\star$ is the reference model~\eqref{eq-selection-m} with lowest perplexity on human texts. This formulation highlights the similarity between \mosaic{} and \bino{} scores, differing only in how they compute average surprisal: a fixed model for \bino{} vs. a position-dependent mixture for \mosaic{}. This score is used to detect artificial texts as follows: given an appropriate threshold $\delta > 0$ and a sample text $\mathbf{y}$, if $S_{m^\star,\mathcal{M}}(\mathbf{y}) \geq \delta$, the text is classified as human; otherwise, it is considered AI-generated.
\end{defn}

A formal description of how this scoring system is implemented is provided  in Algorithm~\ref{alg:mosaic}.

\section{Experimental Settings \label{sec:experimental-settings}}
\subsection{Datasets \& Metrics \label{ssec:datasets}}
\begin{table}[hbt]
    \centering
    \small
    \setlength{\tabcolsep}{1pt}
        \begin{tabular}{l|r|r|r|r|r}
         && \multicolumn{2}{c|}{Human} & \multicolumn{2}{c}{Artificial} \\ 
         Corpus Name      & \multicolumn{1}{c|}{\# Gen}               & \# texts & avg len & \# texts & avg len \\ \hline
         RAID (sampling)  & 11           & 1k  & 452   &  11k  & 373 \\
         RAID adversarial & 11          & 1k  & 452   &  11k  & 658 \\
         RAID+            & 2 & 1k  & 452   &  2k   & 410 \\
         M4 (Multilingual)& 1 & 15k & 729   &  15k  & 649 \\
    \end{tabular}
    \caption{Dataset details: RAID+ was specifically generated for this study using the models considered in our ensembling experiments. \# Gen indicates the numbers of generators used for the artificial texts, avg len represents the average length of texts in \llama-2 tokens.}
    \label{tab:datasets-details}
\end{table}

\begin{figure*}[t!]
    \centering
    \includegraphics[width=\linewidth]{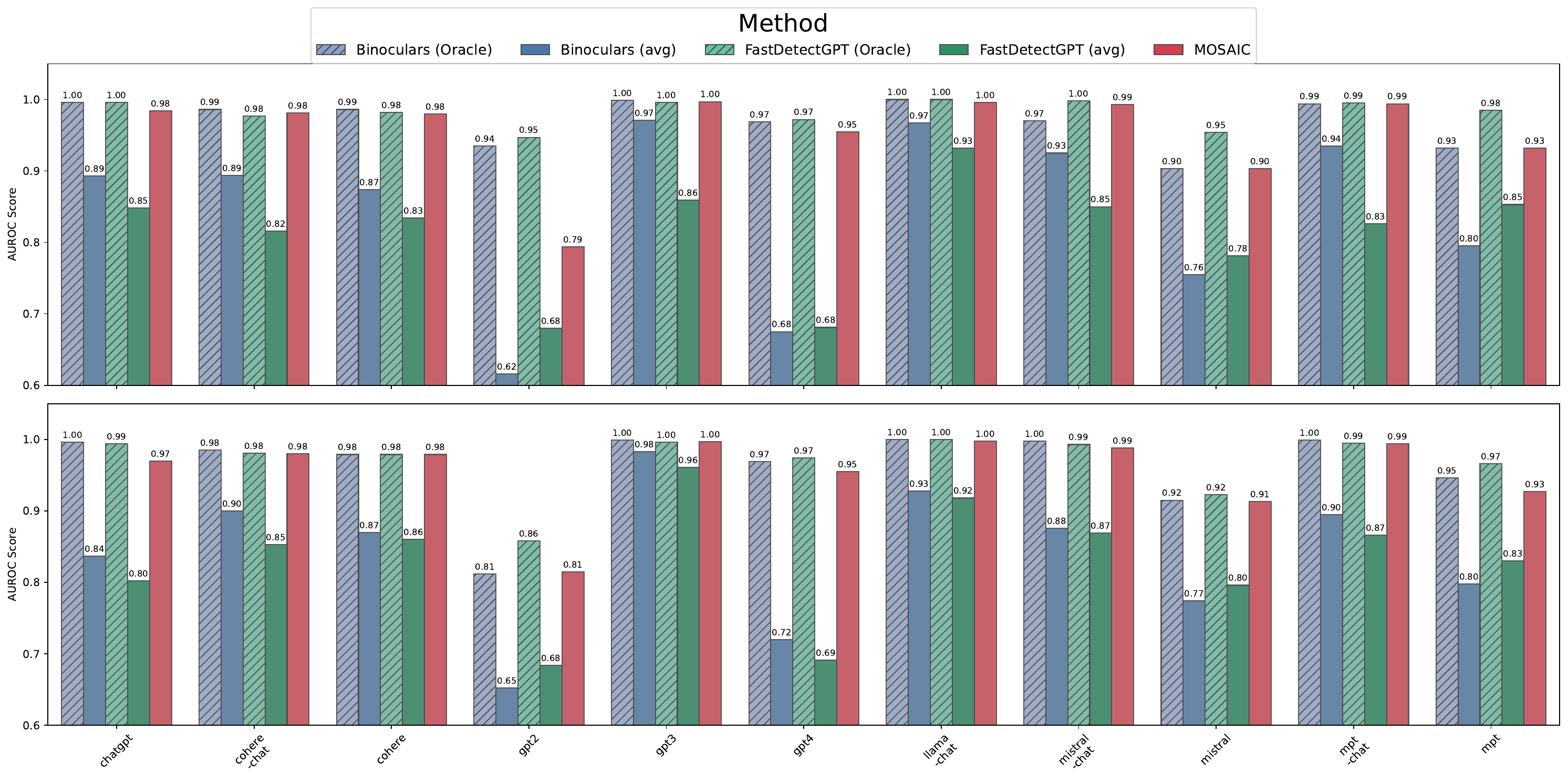}
    \caption{Comparing \mosaic{} AUROC on the RAID dataset with different methods using the \falcon{} family (top plot) and \llama{} and \tower{} models (bottom plot). Model names represent the LLM used to generate the dataset.}
    \label{fig:raid-mosaic}
\end{figure*}

We evaluate our method on a diverse set of texts and generative models from the literature: 
RAID \citep{dugan-etal-2024-raid} and M4 \citep{wang-etal-2024-m4}.

\textbf{RAID} contains about 15k natural texts in English from a variety of domains; the artificial part version contains approximately 500k, generated with a diverse set of recent models, also varying the sampling procedure. As the test set is not publicly released, we select a random subset of 1,000 human texts and their generated counterparts with 11 different models and ancestral sampling for our experiments. RAID also includes an artificially noised subcorpus, we report the results for the different adversarial attacks in Table~\ref{tab:mosaic_robustness} in the Appendix. 

\textbf{The M4\footnote{For \emph{Multi-Lingual, Multi-Domain, Multi-Generator Machine-Generated text}.} corpus} is a massive dataset of natural texts collected from a diverse set of sources \citep{wang-etal-2024-m4}. Comparable artificial texts are generated by 6~LLMs, with prompts such as article titles, headlines, or abstracts depending on their domain. In our experiments, we only use one multilingual generator (ChatGPT, \url{https://chatgpt.com/}), and the ``balanced'' sets made of 3,000 pairs of (artificial, natural) texts in Chinese, Russian, Bulgarian, Arabic, and German.

We augmented the \textbf{RAID} extracts with texts generated using models that are in our ensembles,\footnote{Generation uses ancestral sampling.} using the same prompts as in the RAID dataset. We refer to this augmented RAID as RAID+.

These datasets, presented in Table~\ref{tab:datasets-details}, represent a large variety of genres, themes, languages, sampling strategies, and generators, allowing us to thoroughly assess our detection strategy in different settings. Using RAID+, we can also evaluate detection performance for texts produced by one of our detectors.

\paragraph{Metrics.} As in most studies, we report the AUROC score as our main evaluation metric. Depending on the application, True Positive Rate (TPR) for a predefined False Positive rate (e.g., 5\%) is also worth looking at and reported. All these scores are obtained using scikit-learn \citep{pedregosa-2011-scikit-learn}.

\subsection{Choice of Models}
In the next experiments, we use two ensembles of four models each, sharing a common vocabulary (65k for Falcon, 32k for Llama-2):
\begin{itemize}
    \item The \falcon{}  family \cite{almazrouei-etal-2023-FalconSeriesOpen} (\falcon-7b, \falcon-7b-instruct, \falcon-40b, \falcon-40b-instruct).
    \item \llama{} \cite{touvron-etal-2023-LLaMAOpenEfficient} model variations (\llama-2-7b, \llama-2-7b-chat, \towerbase-7b and \towerbase-13b \cite{alves-etal-2024-tower}).
\end{itemize}
Both ensembles contain LLMs used in the original \bino{} paper, to which we added models to deepen the analysis of our ensembling techniques. We only consider pairs or ensembles of models within one group, as computing the cross-entropy term in equations~\eqref{eq:binoculars-score} and \eqref{eq:mosaic_score} requires models to share the same tokenizer.

\section{Experimental Results \label{sec:results}}
\begin{table*}[t]
    \centering
    \scalebox{0.95}{%
    \setlength{\tabcolsep}{2pt}
    \begin{tabular}{|l|c|c|c|c|c|c|c|c|c|c|c|}
        \hline
        \textbf{AUROC} & \textbf{chatgpt} & \textbf{cohere-c} & \textbf{cohere} & \textbf{gpt2} & \textbf{gpt3} & \textbf{gpt4} & \textbf{llama-c} & \textbf{mistral-c} & \textbf{mistral} & \textbf{mpt-c} & \textbf{mpt} \\
        \hline
        \mosaic-2 & 0.928 & 0.978 & 0.968 & 0.803 & 0.994 & 0.902 & 0.998 & 0.970 & 0.895 & 0.990 & 0.920 \\
        \hline
        \mosaic-3 (i)& 0.917 & 0.978 & 0.971 & 0.818 & 0.994 & 0.898 & 0.996 & 0.969 & 0.903 & 0.988 & 0.925 \\
        \hline
        \mosaic-3 (ii) & 0.972 & 0.980 & 0.979 & 0.802 & 0.997 & 0.956 & 0.998 & 0.987 & 0.909 & 0.994 & 0.922 \\
        \hline
        \mosaic-4 & 0.970 & 0.979 & 0.980 & 0.815 & 0.997 & 0.955 & 0.998 & 0.988 & 0.914 & 0.994 & 0.927 \\
        \hline
        \textbf{TPR@5\%FPR} & \textbf{chatgpt} & \textbf{cohere-c} & \textbf{cohere} & \textbf{gpt2} & \textbf{gpt3} & \textbf{gpt4} & \textbf{llama-c} & \textbf{mistral-c} & \textbf{mistral} & \textbf{mpt-c} & \textbf{mpt} \\
        \hline
        \mosaic-2 & 0.674 & 0.923 & 0.893 & 0.365 & 0.984 & 0.636 & 0.999 & 0.881 & 0.577 & 0.960 & 0.643 \\
        \hline
        \mosaic-3 (i)& 0.629 & 0.922 & 0.904 & 0.377 & 0.984 & 0.610 & 0.998 & 0.871 & 0.584 & 0.948 & 0.644 \\
        \hline
        \mosaic-3 (ii) & 0.844 & 0.937 & 0.924 & 0.314 & 0.990 & 0.769 & 1.000 & 0.938 & 0.546 & 0.978 & 0.586 \\
        \hline
        \mosaic-4 & 0.831 & 0.936 & 0.928 & 0.351 & 0.991 & 0.771 & 1.000 & 0.943 & 0.568 & 0.976 & 0.624 \\
        \hline
    \end{tabular}}
    \caption{Performance of \mosaic{} on the RAID dataset with a varying number of underlying models. ``-c'' indicates the chat version of a model. 2 models: \llama-2-7b and its chat version, 3 models (i): both \llama + \towerbase-7B, 3 models (ii): both \llama + \towerbase-13B, 4 models: all 4 Llama and Tower models.}
    \label{tab:mosaic-2-3-4-models}
\end{table*}
\subsection{Results on RAID}
In Table~\ref{tab:auc-falcon-min-max}, we perform a systematic search to figure out the ``best'' performing configuration for each generator model used in RAID, that we call ``Oracle'' in Figure~\ref{fig:raid-mosaic}. The optimal model selection varies by dataset, making it incompatible with a ``generator-agnostic'' approach. As the average \bino{} and \fast{} scores across combinations indicate, most settings yield poor performance, highlighting the need for a robust criterion to identify effective model combinations.


In Figure~\ref{fig:raid-mosaic}, we observe that \mosaic \textbf{performs as well as the other methods’ oracle configurations}, with the exception of GPT2 generations. This can be explained by the poorer quality of the outputs, increasing the surprisal of artificial texts, thus misleading the detectors. The poor average performance across all combinations on the data generated by this smaller model supports this argument. In contrast, most static two-model combinations struggle to detect GPT4 accurately. However, since GPT4’s outputs are of very high quality, \mosaic{} remains effective in identifying them.


\subsubsection{Augmenting the Ensemble}
To demonstrate the effectiveness of ensembling, we showcase the results of \mosaic{} using 2, 3 and 4 models. We first look at the results of the \llama-2-7b and \llama-2-7b-chat pair, then add \towerbase-7B-v0.1, then \towerbase-13B-v0.1. Following our criterion of ``lowest surprisal over the human texts'' defined in Section~\ref{ssec:formulas}, the model used to compute conditional entropy for the 2 and 3~model setting is \llama-2-7b, and \towerbase-13B-v0.1 is used for the 4-model setting.

In Table~\ref{tab:mosaic-2-3-4-models}, we see that adding the ``best'' model, \towerbase-13B, leads to improved performances, with the exception of ``mistral'' and ``mpt'' generations. Adding the 7B version of \tower{} does not change much the results of our method, probably because its capabilities in English are similar to those of the already available \llama{} models. Augmenting the ensemble seems to have the least effect when the generator model is small, as GPT2, Mistral and MPT are the ones showing few improvements when adding models, even worsening results at times.

\subsection{Including the Generator in the Ensemble}
With our 1,000 human samples of RAID and the prompts used, we generated 1,000 texts using ancestral sampling with both \falcon-40b and \llama-2-7b, then used \mosaic{} with respectively the \falcon{} and the \llama{} families. \mosaic{} with \falcon{} models obtains an AUC of 0.965 for the text generated by \falcon-40b, and \mosaic{} with the \llama{} and \tower{} models obtains an AUC of 0.986 for the texts generated by \llama-2-7b. These results are good but surprisingly average for the methods, suggesting that the size of the generator and the quality of its outputs matter more than its usage as a detector. In our setting, having the generator in the ensemble does not appear particularly advantageous. 

\subsection{Choosing the Best ``Reference'' Model \label{ssec:choosing-m}}

As mentioned in Section~\ref{ssec:formulas}, the model $m$ used to compute the conditional entropy needs to be the one with the least surprisal when looking at human texts. When having a family of models available, the one with the most parameters often ends up being better at this task and becomes the obvious choice. However, we purposely chose the \tower{} models for their multilingual capabilities, despite a number of parameters similar to the \llama{} models. Table~\ref{tab:log-perplexity} reports the log-perplexities of these models on the ``human'' parts of the M4 dataset. Table~\ref{tab:different-m-auroc} then gives us the performance of \mosaic{} when using different $m$ models to compute the conditional entropy. Our heuristic of selecting the reference model based on its low perplexity on human texts (see Section~\ref{ssec:formulas}) consistently gives the best results. This simple criterion only requires human development texts; if unavailable, the largest model in the ensemble can serve as a good proxy.


\begin{table}[ht]
    \centering
    \small
    \setlength{\tabcolsep}{1pt}
    \begin{tabular}{|l|c|c|c|c|c|}
        \hline
        & Arabic & Bulgarian & Chinese & German & Russian \\
        \hline
        TowerBase-13B   & \textbf{1.2743}    & \textbf{1.8052}    & \underline{2.3047} & \textbf{1.4912}    & \textbf{1.5069}    \\ \hline
        TowerBase-7B    & 1.3929             & 1.9839             & 2.3527             & 1.6169             & \underline{1.6084} \\ \hline
        Llama-2-7b-chat & 1.7379             & 2.3175             & 2.6800             & 2.1189             & 2.2917             \\ \hline
        Llama-2-7b      & \underline{1.3506} & \underline{1.8291} & \textbf{2.1286}    & \underline{1.6117} & 1.7778             \\ \hline
    \end{tabular}
    \caption{Log-perplexity values of our models for the ``human'' texts in M4}
    \label{tab:log-perplexity}
\end{table}

\begin{table}[ht]
    \centering
    \small
    \setlength{\tabcolsep}{1pt}
    \begin{tabular}{|l|c|c|c|c|c|}
        \hline
        Model $m$       & Arabic             & Bulgarian          & Chinese            & German             & Russian \\
        \hline
        TowerBase-13B   & \textbf{0.9563}    & \textbf{0.9888}    & \textbf{0.9752}    & \textbf{0.9311}    & \textbf{0.9148}    \\ \hline
        TowerBase-7B    & \underline{0.9111} & 0.9578             & \underline{0.9558} & 0.8679             & \underline{0.8569} \\ \hline
        Llama-2-7b-chat & 0.7768             & 0.8262             & 0.5849             & 0.6751             & 0.4321             \\ \hline
        Llama-2-7b      & 0.8947             & \underline{0.9762} & 0.9059             & \underline{0.9200} & 0.6814             \\ \hline
    \end{tabular}
    \caption{AUROC of \mosaic{} on the M4 dataset when varying the ``reference'' model $m^\star$.}
    \label{tab:different-m-auroc}
\end{table}

\subsection{Using Multiple Variants of a Single Model \label{ssec:one-model-mode}}
A lighter and faster implementation of \mosaic{} can be developed based on one single model, the logits of which are modified in order to simulate several probability distributions, that can then be ensembled. In fact, many generation techniques are based on the adaptation of the distribution of an underlying model \citep{meister-etal-2023-efficacy}, such as top-k sampling \citep{fan-etal-2018-hierarchical}, top-p or nucleus sampling \citep{holtzman-etal-2020-thecurious},  $\eta$-sampling \cite{hewitt-etal-2022-truncation}, etc. Using such techniques in \mosaic{} only requires to load in memory and perform inference with just one model; it also readily satisfies the constraint that all distributions in the ensemble should share a common vocabulary.

We explored this with \llama-2-7b and four different values of nucleus sampling.\footnote{In our implementation, we use a smoothed version of the adapted distribution, so that all tokens have a small probability to be sampled. Having the same support for all distributions is required to compute the cross-entropy term in Eq.~\eqref{eq:mosaic_score}.} Results are reported in the Appendix (Table~\ref{tab:one_model_mosaic_auc}). This choice leads to results that are weaker than the 2-models setting depicted in Table~\ref{tab:mosaic-2-3-4-models}, suggesting that applying this top-p in four different ways does not introduce as much diversity as the instruct model does. Another downfall of this approach is that the selection of $m$, the ``reference model'', can no longer rely on perplexity scores, as these cannot be reliably computed when with truncated vocabularies. In our experiment, we thus experimented with all potential values. Further work is needed to draw more precise analyses of this use-case.

\subsection{Robustness Issues \label{ssec:robustness}}
Using the same 1000 human samples of RAID considered in all our previous experiments, we randomly sampled 1000 artificially generated texts for each adversarial attack available in RAID. Performance on this ``noised'' test set are in Table~\ref{tab:mosaic_robustness} in the Appendix. \mosaic{} (with \llama{} and \tower{}) is quite resilient to such modifications, with the exception of ``synonym'' and ``zero-width space'' attacks, which significantly deteriorate the performance of the method due to the large change in surprisal they induce in the generated texts.

\subsection{Uniform Ensembling \label{ssec:uniform-mosaic}}
Instead of using the Blahut-Arimoto algorithm for optimal model combination, we naively assigned equal weights to each model in the ensemble. These results are reported in Table~\ref{tab:mosaic-raid-falcon-llama} on row ``\mosaic{} (uniform \falcon{})''. We observe that this uniform mixture yields good overall results; yet, it still underperforms our theory-driven ensembling method for every generator in the RAID dataset.

\section{Conclusion \label{sec:conclusion}}
Through all of our experiments, we have shown that the \mosaic{} method effectively combines all the models in the ensemble, achieving very strong results across all datasets and languages. This method has multiple advantages: it is fully unsupervised, dispenses with the search for the optimal detector(s) when several are available, while offering a scalable solution that can incorporate a growing number of models. Even adversarial attacks have only a minimal effect on the performance, despite our scoring system not being optimized for perturbations. 
However, \mosaic{} is currently computationally costly, as each model must run on the input text. 

While not in the main scope on this paper, a potential improvement of computational efficiency is proposed in Section~\ref{ssec:one-model-mode}, and further solutions are discussed in Appendix~\ref{ann:compute improvements}. Furthermore, more work needs to be done in order to fully understand how to evaluate the ``distance'' in-between models’ outputs, in order to choose the best ensemble that would cover all potential generations.

The optimal mixture in \mosaic{}, defined by Blahut-Arimoto weights to minimize encoding length in worst-case scenarios, could be useful beyond detecting machine-generated texts. We leave this exploration for future work.


\section{Limitations \label{sec:limitations}}

As mentioned above, the computational cost of the method, running multiple LLMs as well as the Blahut-Arimoto algorithm for each token is currently an issue, being able to analyse the 1000 texts of each RAID sub-dataset with 4 models in about an hour on one A100 GPU if using the \llama{} and \tower{} models, and three if using the \falcon{} ones. 
Theory-wise, the main issue with our method is the need to compute the cross-entropy between models, forcing them to have the same vocabulary, thus greatly limiting the number of models we can combine at once. 

\section{Ethical Considerations}
It should be acknowledged that artificial text detection tools are not infallible and consequently should not be used as the sole basis for punitive actions or decisions that could affect individuals or organizations. These methods must be complemented by human oversight and verification before taking any drastic measure, to ensure fairness of treatment.


\section*{Acknowledgments}
This work was performed using HPC resources from GENCI–IDRIS (Grant 2023-AD011014903).

\bibliography{anthology,custom}

\appendix
\section{Information-Theoretic Principles Behind FastDetectGPT and Binoculars \label{ann:theory-similarities}} 
As discussed in Section~\ref{ssec:formulas}, \fast{} closely resembles \bino{}. However, rather than directly computing the cross-entropy between two models, it draws $N$ independent samples from the model and computes the empirical cross-entropy:
$ \{\tilde{y}_i \sim p(Y_t | \mathbf{y}_{<t})\}_{i=1}^N$ and then return the  score:
\begin{equation*}
    S_{p,q}^{\textrm{Fast}}(\mathbf{y})= \frac{-\log q(y_t | \mathbf{y}_{<t}) + \frac{1}{N} \sum\limits_{i=1}^N \log q(\tilde{y}_i | \mathbf{y}_{<t})}{\tilde{\sigma}(\mathbf{y}_{<t})},
\end{equation*}
where 
\begin{align}
\tilde{\sigma}^2 (\mathbf{y}_{<t}) & \triangleq \frac{1}{N-1} \sum\limits_{i=1}^N \Big(-\log q(y_i | \mathbf{y}_{<t})   \nonumber\\
&+  \frac{1}{N} \sum_{j=1}^N \log q(y_j | \mathbf{y}_{<t}) \Big)^2
\end{align}
is a normalisation term which is particularly useful for individual token detection.

\section{Theoretical Grounding of \mosaic{}}
\textbf{Identifying explanations of data.}  We turn to the problem of determining an adequate sequence of models  $\hat{\mathbf{m}}= \langle \hat{m}_0, \ldots, \hat{m}_{T} \rangle$. 

Our goal will be to derive a robust scoring algorithm that best extracts regularity in the data, which is equivalent to identifying  \textbf{the model that achieves the best compression of the input tokens}. Suppose we are given a family of LLMs $\mathcal{P}_{\mathcal{M}}(\mathcal{Y})$, with corresponding Shannon codelengths: 
$$
\mathcal{L}_{p_m}(y_t |\mathbf{y}_{<t})\triangleq -\log p_m ( y_t | \mathbf{y}_{<t}), 
$$ 
for each $y_t$. These can be viewed as a collection of data compressors, indexed by $m$. We can measure the performance of encoding $y_t$ at time $t$ relative to $\mathcal{P}_{\mathcal{M}}(\mathcal{Y})$. If we choose to encode the token $y_t$ with model $q ( y_t | \mathbf{y}_{<t})$, the resulting expected excess codelength (or overhead) w.r.t. any distribution $p_m \in \mathcal{P}_{\mathcal{M}}(\mathcal{Y})$ is: \vspace{1mm}
\begin{align}
&\mathcal{R}_m(q\,  ; \mathbf{y}_{<t}) \triangleq   \underset{y_t \sim p_m ( y_t |\mathbf{y}_{<t})}{ \mathbb{E}}\Big[-\log q ( y_t | \mathbf{y}_{<t})\Big] \notag \\ & \hspace{10em} - \mathcal{H}_{p_m}(Y_t| \mathbf{y}_{<t})\nonumber \vspace{1mm}
\end{align}
which is non-negative since $\mathcal{H}_{p_m}(Y_t|\mathbf{y}_{<t})$ is the \emph{minimum  expected codelength}. $\mathcal{R}_{m}$ represents the extra averaged number of bits needed to encode $y_t$ using the code/LLM $q ( y_t | \mathbf{y}_{<t})$, as compared to $\mathcal{H}_{p_m}(Y_t| \mathbf{y}_{<t})$, the number of bits needed if we would have used the best fitting LLM in $\mathcal{P}_{\mathcal{M}}(\mathcal{Y})$  with hindsight. However, the encoder cannot know the underlying model artificially generating $y_t$ so we  take a worst-case approach and look for universal LLMs with small worst-case expected overhead, where the worst-case is over all models in $\mathcal{P}_{\mathcal{M}}(\mathcal{Y})$. $ \mathcal{R}_{m}$ is our quality measure and hence, the `optimal' LLM relative to $\mathcal{P}_{\mathcal{M}}(\mathcal{Y})$, for a given context $\mathbf{y}_{<t}$,  is the distribution minimizing: \vspace{1mm}
\begin{equation}
q^\star( y_t | \mathbf{y}_{<t}) \triangleq   \argmin_{q \in \mathcal{P}(\Omega)} \, \max_{m \in\mathcal{M}} \mathcal{R}_{m}(q\,;\mathbf{y}_{<t}) \vspace{1mm}   \label{eq-optimal}
\end{equation}
where the minimum is over all distributions on $\Omega$. The minimizer corresponds to the code with the smallest overhead (i.e., the fewest extra bits) relative to the optimal code, which is retrospectively the best choice in the worst-case model selection for generating synthetic text across all LLMs in the available family $ \mathcal{P}_{\mathcal{M}}(\mathcal{Y})$. 
\vspace{1mm}

\textbf{Leveraging codelengths for identifying AI-generated text.}  The averaged overhead of the optimal codelength $-\log q^\star( y_t | \mathbf{y}_{<t}) $ obtained by solving Eq.~\eqref{eq-optimal} seems to be a very reasonable choice for building a robust score function to detect AI-generated text because of the following properties: 
\begin{itemize}
    \item The better the best-fitting LLM in $\mathcal{P}_{\mathcal{M}}(\mathcal{Y})$  fits the artificially generated data, the shorter the codelengh $\mathcal{L}_{q^\star}(y_t | \mathbf{y}_{<t}) \triangleq -\log q^\star( y_t | \mathbf{y}_{<t}) $.
    \item No LLM in $\mathcal{P}_{\mathcal{M}}(\mathcal{Y})$ is given a prior preference over any other since $\mathcal{R}_{m}(q^\star\,  ; \mathbf{y}_{<t}) $ $  \leq\mathcal{R}_{m}(p \,  ; \mathbf{y}_{<t}) $ for all $p \in \mathcal{P}_{\mathcal{M}}(\mathcal{Y})$, i.e., we are treating all LLMs within our universe  $\mathcal{P}_{\mathcal{M}}(\mathcal{Y})$ on the same footing. 
\end{itemize}
\subsection{Proof of Proposition~\ref{proposition}   \label{ann:theory}}
\begin{proof}
We need to show the fundamental identity:     
\begin{align}
\Gamma(\mathbf{y}_{<t}) & \triangleq  \min_{q \in \mathcal{P}(\Omega) } \max_{m\in\mathcal{M}} \mathcal{R}_{m}(q\,;\mathbf{y}_{<t})  \\
 &= \max_{ \mu\in \mathcal{P}(\mathcal{M})  }\,  \mathcal{I}(\mathbb{M} ; Y_t | \mathbf{y}_{<t} ), \label{eq-to-show}
  \end{align}
  where the optimal $q^\star( y_t | \mathbf{y}_{<t})$ achieving the minimum is characterized by the mixture: 
\begin{equation}
q^\star( y_t | \mathbf{y}_{<t}) = \sum\limits_{m\in\mathcal{M}} \mu^\star(m|\mathbf{y}_{<t} ) p_m ( y_t | \mathbf{y}_{<t})  
\end{equation}
and the distribution $\mu^\star(m|\mathbf{y}_{<t})$ of the random variable $\mathbb{M}$ on $\mathcal{M}$ follows by solving:  
\begin{equation}
\mu^\star(m|\mathbf{y}_{<t}) \triangleq  \argmax_{ \mu \in  \mathcal{P}(\Omega) }\,  \mathcal{I}\big(\mathbb{M} ; Y_t | \mathbf{y}_{<t} \big). \label{eq-last}
\end{equation}
To this end, we start from the definition  $\mathcal{R}_{m}$: 
\begin{align}
\mathcal{R}_{m}(q\,  ; \mathbf{y}_{<t}) \triangleq  
& \underset{y_t \sim p_m ( y_t |\mathbf{y}_{<t})}{ \mathbb{E}}\left[- \log q ( y_t | \mathbf{y}_{<t})\right]  \notag \\ \notag
& \hspace{-6em} -\min_{q^\prime\in\mathcal{P}(\Omega)} \, \underset{y_t \sim p_m ( y_t | \mathbf{y}_{<t})}{\mathbb{E}}\Big[-\log q^\prime ( y_t | \mathbf{y}_{<t}) \Big]  \\ \notag
& \hspace{-7.5em} = \underset{y_t \sim p_m ( y_t | \mathbf{y}_{<t})}{ \mathbb{E}}\left[-\log q ( y_t | \mathbf{y}_{<t})\right]  - \mathcal{H}_{p_m} (Y_t| \mathbf{y}_{<t})  \\ 
 & \hspace{-7.5em} = \mathcal{D}_{\textrm{KL}}\Big( p_m ( \cdot | \mathbf{y}_{<t}) \big \| q ( \cdot | \mathbf{y}_{<t})  \Big), 
\end{align}
where $\mathcal{D}_{\textrm{KL}}(\cdot\| \cdot)$ denotes the Kullback–Leibler divergence. Hence, we can formally state our problem as follows: 
\begin{align}
 \Gamma(\mathbf{y}_{<t}) &= \min_{q \in \mathcal{P}(\Omega) } \max_{m\in\mathcal{M}} \mathcal{R}_{m}(q\,;\mathbf{y}_{<t})  \nonumber\\
& \hspace{-3em} =\min_{q \in \mathcal{P}(\Omega) } \max_{m\in\mathcal{M}}  \mathcal{D}_{\textrm{KL}}\Big( p_m( \cdot| \mathbf{y}_{<t}) \big \| q ( \cdot | \mathbf{y}_{<t})  \Big) \nonumber \\
& \hspace{-3em} = \min_{q \in \mathcal{P}(\Omega) } \max_{\mu \in\mathcal{P}(\mathcal{M})}  \underset{m\sim \mu}{\mathbb{E}} \mathcal{D}_{\textrm{KL}}\Big( p_m ( \cdot | \mathbf{y}_{<t}) \big \| q ( \cdot | \mathbf{y}_{<t})  \Big), \label{eq-KL}
\end{align}
where the minimum is taken over all the possible distributions $q \in \mathcal{P}(\Omega)$, representing the expected value of
regret of $q$ w.r.t. the worst-case distribution over $\mu \in\mathcal{P}(\mathcal{M})$. Notice that this  is  equivalent to the \emph{average worst-case regret}~\cite{BarronRY1998TInfT, 9646872}. The equality in \eqref{eq-KL} holds by noticing that 
\begin{align}
   \max_{\mu \in\mathcal{P}(\mathcal{M})} \,  \underset{m\sim \mu}{\mathbb{E}} \mathcal{D}_{\textrm{KL}}\Big( p_m ( \cdot | \mathbf{y}_{<t}) \big \| q ( \cdot | \mathbf{y}_{<t})  \Big) \notag \\ 
    & \hspace{-15em} \leq \max_{m\in \mathcal{M}}  \mathcal{D}_{\textrm{KL}}\Big( p_m( \cdot | \mathbf{y}_{<t}) \big \| q ( \cdot | \mathbf{y}_{<t})  \Big)  
\end{align}
and moreover, 
\begin{align}
\max_{m\in \mathcal{M}}   \, \mathcal{D}_{\textrm{KL}}\Big( p_m ( \cdot | \mathbf{y}_{<t}) \big \| q ( \cdot | \mathbf{y}_{<t})  \Big) \notag \\   
    & \hspace{-14em} = \underset{m\sim \widetilde{\mu}}{\mathbb{E}} \mathcal{D}_{\textrm{KL}}\Big( p_m( \cdot | \mathbf{y}_{<t}) \big \| q ( \cdot | \mathbf{y}_{<t})  \Big)
  \end{align}  
by choosing the measure $\widetilde{\mu}$ to be an uniform probability over the set $\widetilde{\mathcal{M}}\subseteq \mathcal{M}$, which is defined as the set of maximizers: 
\begin{align*}
\widetilde{\mathcal{M}} \equiv  \argmax_{m\in \mathcal{M}}  \mathcal{D}_{\textrm{KL}}\Big( p_m( \cdot | \mathbf{y}_{<t}) \big \| q ( \cdot | \mathbf{y}_{<t})  \Big) 
 \end{align*}  
and zero otherwise. 

The convexity of the KL-divergence allows us to rewrite expression~\eqref{eq-KL} as follows:
\begin{align}
&\min_{q \in \mathcal{P}(\Omega) } \max_{\mu \in\mathcal{P}(\mathcal{M})}  \underset{m\sim \mu}{\mathbb{E}} \mathcal{D}_{\textrm{KL}}\Big( p_m ( \cdot | \mathbf{y}_{<t}) \big \| q ( \cdot | \mathbf{y}_{<t})  \Big)   \notag \\
 = &  \max_{\mu \in\mathcal{P}(\mathcal{M})}  \min_{q \in \mathcal{P}(\Omega) } \underset{m\sim \mu}{\mathbb{E}}  \mathcal{D}_{\textrm{KL}}\Big( p_\theta ( \cdot | \mathbf{y}_{<t}) \big \| q ( \cdot | \mathbf{y}_{<t})  \Big) \label{eq:minimaxProb3}
   \end{align}  
This follows by considering a zero-sum game with a concave-convex mapping defined on a product of convex sets. The sets of all probability distributions $\mathcal{P}(\mathcal{M})$ and $\mathcal{P}(\Omega)$
 are two nonempty convex sets, bounded and finite-dimensional. On the other hand, $ (\mu , q ) \rightarrow  \underset{m\sim \mu}{\mathbb{E}}  \mathcal{D}_{\textrm{KL}}\Big( p_m ( \cdot | \mathbf{y}_{<t}) \big \| q ( \cdot | \mathbf{y}_{<t})  \Big)$  is a concave-convex mapping, i.e., 
  $$\mu \rightarrow  \underset{m\sim \mu}{\mathbb{E}}  \mathcal{D}_{\textrm{KL}}\Big( p_m( \cdot |\mathbf{y}_{<t}) \big \| q ( \cdot | \mathbf{y}_{<t})  \Big) $$ 
 is concave and, $$q \rightarrow   \underset{m\sim \mu}{\mathbb{E}}  \mathcal{D}_{\textrm{KL}}\Big( p_m ( \cdot | \mathbf{y}_{<t}) \big \| q ( \cdot | \mathbf{y}_{<t})  \Big) $$ is convex for every $ (\mu , q )$, respectively.  Then, by classical min-max theorem~\cite{vonNeumann1928-VONZTD-2}, we have that~\eqref{eq:minimaxProb3} holds. 

Finally, it remains  to show that: 
\begin{align}
\min_{q \in \mathcal{P}(\Omega)} \underset{m\sim \mu}{\mathbb{E}} & \, \mathcal{D}_{\textrm{KL}}\Big( p_m ( \cdot | \mathbf{y}_{<t}) \big \| q ( \cdot | \mathbf{y}_{<t})  \Big) \notag \\ & =  \mathcal{I}(\mathbb{M} ; Y_t | \mathbf{y}_{<t} ) \label{eq-missing-appex}
    \end{align}
for any random variable $\mathbb{M}$ distributed according to the probability distribution $\mu \in  \mathcal{P}(\mathcal{M})$ and each distribution $ p_m ( y_t | \mathbf{y}_{<t})$. 

We begin by showing that: 
\begin{align*}
\underset{m\sim \mu}{\mathbb{E}} & \mathcal{D}_{\textrm{KL}}\Big( p_m ( \cdot | \mathbf{y}_{<t}) \big \| q ( \cdot | \mathbf{y}_{<t})  \Big) \geq \mathcal{I}(\mathbb{M} ; Y_t | \mathbf{y}_{<t} )
\end{align*}
for all distributions $q ( \cdot | \mathbf{y}_{<t})  $ and   $ p_m ( y_t | \mathbf{y}_{<t})$. To this end, we consider  the following identities: 
\begin{align}
\underset{m\sim \mu}{\mathbb{E}}  \mathcal{D}_{\textrm{KL}}\Big( p_m( \cdot | \mathbf{y}_{<t}) \big \| q ( \cdot | \mathbf{y}_{<t})  \Big) \notag \\
& \hspace{-13em}  =\underset{m\sim \mu}{\mathbb{E}}  \mathcal{D}_{\textrm{KL}}\Big( p_m ( \cdot | \mathbf{y}_{<t}) \big \| p_m ( \cdot | \mathbf{y}_{<t}) \Big) \nonumber \\
& \hspace{-10em}+ \mathcal{D}_{\textrm{KL}}\Big(  p_m ( \cdot | \mathbf{y}_{<t})  \big \| q ( \cdot | \mathbf{y}_{<t}) \Big) \nonumber \\
& \hspace{-13em}= \mathcal{I}(\mathbb{M} ; Y_t | \mathbf{y}_{<t} ) + \mathcal{D}_{\textrm{KL}}\Big(  p_m ( \cdot | \mathbf{y}_{<t})  \big \| q ( \cdot | \mathbf{y}_{<t}) \Big)\nonumber \\
& \hspace{-13em} \geq \mathcal{I}(\mathbb{M} ; Y_t | \mathbf{y}_{<t} ), \label{eq-missing-appex-B}
\end{align}
where $ p_m ( \cdot | \mathbf{y}_{<t})$ denotes the marginal distribution of $p_m ( \cdot | \mathbf{y}_{<t})$ w.r.t. $\mu$ and the last inequality follows since the KL divergence is non-negative. Finally, it is easy to check that by selecting:  
\begin{equation} \label{eq-missing-eq}
 q^\star ( y_t | \mathbf{y}_{<t}) =    \underset{m\sim \mu}{\mathbb{E}}  \big[p_m ( y_t | \mathbf{y}_{<t}) \big] 
\end{equation}
the lower bound in \eqref{eq-missing-appex-B} is achieved: 
\begin{align}
\min_{q \in \mathcal{P}(\Omega)} \underset{m\sim \mu}{\mathbb{E}}  \, \mathcal{D}_{\textrm{KL}}\Big( p_m ( \cdot | \mathbf{y}_{<t}) \big \| q ( \cdot | \mathbf{y}_{<t})  \Big) \\
& \hspace{-13em} = \underset{m\sim \mu}{\mathbb{E}}  \, \mathcal{D}_{\textrm{KL}}\Big( p_m ( \cdot | \mathbf{y}_{<t}) \big \| q^\star ( \cdot | \mathbf{y}_{<t})  \Big)
\end{align}
for every  $\mu\in\mathcal{P}(\mathcal{M})$, which proves the identity in expression \eqref{eq-missing-appex}. 

The claim in \eqref{eq-to-show} follows by taking the maximum overall probability measures  $\mu \in \mathcal{P}(\mathcal{M})$ at both sides of \eqref{eq-missing-appex}, and combining the resulting identity with expressions  \eqref{eq:minimaxProb3} and \eqref{eq-KL}. The mixture in \eqref{eq-last} follows from expression \eqref{eq-missing-eq} which is a necessary condition to solve the min-max problem. 
\end{proof}

\section{Blahut–Arimoto Algorithm~\label{ann:Blahut–Arimoto}}


\subsection{Algorithm}
Our channel can be specified using two discrete random variables \( (\mathbb{M}, Y_t) \) with alphabets \((\mathcal{M}, \Omega) \) and  probability distributions $\mu$ and  $p_m ( y_t | \mathbf{y}_{<t})$, respectively,  conditioned on $\mathbf{y}_{<t}$. The problem to be solved is the maximization of the mutual information, which consists in:
\begin{equation}
\Gamma(\mathbf{y}_{<t}) \triangleq  \max_{ \mu\in\mathcal{P}(\mathcal{M})}\,  \mathcal{I}_m \big(\mathbb{M} ; Y_t | \mathbf{y}_{<t} \big) .    \label{eq-arimoto}
\end{equation}
Now if we denote the cardinality \(|\mathbb{M}| = M\), \(|\Omega| = N\), then $p_m( y_t | \mathbf{y}_{<t})$ is an \(M \times N \) matrix, which we denote the \( i \)-th row, \( j \)-th column entry by \( w_{ij} \). For the case of channel capacity, the algorithm was introduced in~\cite{Blahut,Arimoto} to solve \eqref{eq-arimoto}. They both found the following expression for the capacity of a discrete channel  with channel law  \( w_{ij} \):
\[
\Gamma(\mathbf{y}_{<t}) = \max_{{\mu}} \max_{Q} \sum_{i=1}^{M} \sum_{j=1}^{N} \mu_{i} w_{ij} \log \left( \frac{q_{ji}}{\mu_{i}} \right),
\]
where \( \mathbf{\mu} \) and \( Q \) are maximized over the following requirements:
\begin{itemize}
    \item \( {\mu} \triangleq (\mu_1,\dots,\mu_M) \) is a probability distribution on \( \mathcal{M} \). That is,  \( \sum_{i=1}^{M} \mu_{i} = 1 \).
    \item \( Q = (q_{ji}) \) is an \( N \times M \) matrix that behaves like a transition matrix from \( \Omega \) to \( \mathcal{M} \) with respect to the channel law. That is, for all \( 1 \leq i \leq M \), \( 1 \leq j \leq N \):
    \[
    q_{ji} \geq 0, \quad q_{ji} = 0 \Leftrightarrow w_{ij} = 0,
    \]
    and every row sums up to 1: 
     $  \sum_{i=1}^{M} q_{ji} = 1. $ 
\end{itemize}
Then, upon initializing a probability measure \( \mathbf{\mu}^0 =(\mu_{1}^0, \mu_{2}^0, \ldots, \mu_{M}^0) \) on \( \mathcal{M} \), we can generate a sequence \( (\mathbf{\mu}^0, Q^0, \mathbf{\mu}^1, Q^1, \ldots) \) iteratively as follows:
\begin{equation}\label{eq-iteration-1}
    (q_{ji}^t) = \frac{\displaystyle \mu_{i}^t w_{ij}}{\displaystyle \sum_{k=1}^{M} \mu_{k}^t w_{kj}}
\end{equation}
and
\begin{equation}\label{eq-iteration-2}
\mu_{k}^{t+1} = \frac{\displaystyle \prod_{j=1}^{N} (q_{jk}^t)^{w_{kj}}}{\displaystyle \sum_{i=1}^{M} \prod_{j=1}^{N} (q_{ji}^t)^{w_{ij}}}
\end{equation}
for \( t = 0, 1, 2, \ldots \).

Then, using the theory of optimization, specifically coordinate descent, it has been shown that the sequence indeed converges to the required maximum. That is,
\[
\lim_{t \to \infty} \sum_{i=1}^{M} \sum_{j=1}^{N} \mu_{i}^t w_{ij} \log \left( \frac{q_{ji}^t}{\mu_{i}^t} \right) = \Gamma(\mathbf{y}_{<t}).
\]
So given a channel law $p_m ( y_t | \mathbf{y}_{<t})$, the \eqref{eq-arimoto} can be numerically estimated up to arbitrary precision.

\subsection{Computational complexity \label{ann:complexity}}

The computational complexity of the Blahut-Arimoto algorithm can be characterized as follows:
\begin{itemize}
    \item  \textbf{Number of iterations.}  The algorithm typically converges linearly, so the number of iterations required, denoted as \(T\), is proportional to the desired accuracy of the solution.
 \item \textbf{Operations per iteration.} Each iteration involves updating the probability measures in \eqref{eq-iteration-1} and \eqref{eq-iteration-2}, and evaluating the mutual information, which requires matrix manipulations. Let $M$ and $N$ be the cardinalities of the input and output alphabets, respectively. Each iteration involves operations over all input-output pairs, requiring \(\mathcal{O}(M \times N)\) operations.
\end{itemize}
Combining these, the overall computational complexity of the Blahut-Arimoto algorithm is \(\mathcal{O}(T \times n \times m)\), reflecting its dependence on the sizes of $M$ (number of LLMs in the considered family) and $N$ (the vocabulary), and the number of iterations needed for convergence, which depends intrinsically on the underlying distributions.

\section{Complementary Results \label{ann:robustness}}
This section is dedicated to additional results. 
Table~\ref{tab:mosaic-raid-falcon-llama} shows the results of \mosaic{} with both ensemble of models on the RAID dataset, along with the uniform mixture discussed in Section~\ref{ssec:uniform-mosaic}.

Table~\ref{tab:one_model_mosaic_auc} reports the results of Section~\ref{ssec:one-model-mode}, using only one model but modifying its logits with nucleus sampling in order to create different probability distributions.

Table~\ref{tab:mosaic_robustness} shows the performance of \mosaic{} under the different adversarial attacks available in the RAID dataset.

\begin{table*}[t]
    \centering
    \small
    \setlength{\tabcolsep}{2pt}
    \begin{tabular}{|l|c|c|c|c|c|c|c|c|c|c|c|}
        \hline
        \textbf{AUROC} & \textbf{chatgpt} & \textbf{cohere-c} & \textbf{cohere} & \textbf{gpt2} & \textbf{gpt3} & \textbf{gpt4} & \textbf{llama-c} & \textbf{mistral-c} & \textbf{mistral} & \textbf{mpt-c} & \textbf{mpt} \\
        \hline
        \mosaic{} (\falcon{} family)        & 0.984 & 0.981 & 0.980 & 0.794 & 0.997 & 0.955 & 0.996 & 0.993 & 0.903 & 0.994 & 0.932 \\
        \hline
        \mosaic{} (uniform Falcon)          & 0.969 & 0.947 & 0.942 & 0.675 & 0.987 & 0.876 & 0.988 & 0.970 & 0.774 & 0.983 & 0.796 \\
        \hline
        \mosaic{} (\llama{} and \tower{})   & 0.970 & 0.980 & 0.979 & 0.815 & 0.997 & 0.955 & 0.998 & 0.988 & 0.913 & 0.994 & 0.927 \\
        \hline
    \end{tabular}
    \caption{Performance of \mosaic{} on the RAID dataset with both families of models.}
    \label{tab:mosaic-raid-falcon-llama}
\end{table*}

\begin{table*}[htbp]
    \centering
    \resizebox{\textwidth}{!}{
    \setlength{\tabcolsep}{2pt}
    \begin{tabular}{l|c|c|c|c|c|c|c|c|c|c|c}
        \toprule
         & chatgpt & cohere-chat & cohere & gpt2 & gpt3 & gpt4 & llama-chat & mistral-chat & mistral & mpt-chat & mpt \\
        \midrule
        p = 0.7 & 0.792 & 0.665 & 0.577 & 0.356 & 0.733 & 0.503 & 0.869 & 0.763 & 0.465 & 0.731 & 0.488 \\
        p = 0.8 & 0.789 & 0.665 & 0.577 & 0.354 & 0.732 & 0.502 & 0.867 & 0.760 & 0.462 & 0.731 & 0.485 \\
        p = 0.9 & 0.781 & 0.663 & 0.578 & 0.358 & 0.725 & 0.503 & 0.860 & 0.752 & 0.457 & 0.726 & 0.480 \\
        p = 0.95& 0.764 & 0.656 & 0.573 & 0.370 & 0.711 & 0.497 & 0.848 & 0.736 & 0.456 & 0.713 & 0.476 \\
        \bottomrule
    \end{tabular}
    }
    \caption{\mosaic{} on RAID using \llama-2-7b with four different values of nucleus sampling, on the RAID dataset. Each row corresponds to the chosen value of p computed as $m$, the ``reference model''.}
    \label{tab:one_model_mosaic_auc}
\end{table*}

\begin{table*}[htbp]
    \centering
    \resizebox{\textwidth}{!}{
    \setlength{\tabcolsep}{2pt}
    \begin{tabular}{l|c|c|c|c|c|c|c|c|c|c|c}
        \toprule
         & \makecell{homoglyph} & \makecell{number} & \makecell{article\\deletion} & \makecell{insert\\paragraphs} & \makecell{misspelling} & \makecell{upper\\lower} & \makecell{whitespace} & \makecell{zero-width\\space} & \makecell{synonym} & \makecell{paraphrase} & \makecell{alternative\\spelling} \\
        \midrule
        AUROC & 0.961 & 0.936 & 0.920 & 0.952 & 0.948 & 0.928 & 0.927 & 0.754 & 0.681 & 0.944 & 0.947 \\
        TPR@5\%FPR   & 0.749 & 0.736 & 0.693 & 0.785 & 0.771 & 0.699 & 0.707 & 0.007 & 0.315 & 0.752 & 0.771 \\
        \bottomrule
    \end{tabular}
    }
    \caption{\mosaic{} AUROC and TPR@5\%FPR for the various attacks performed on RAID, the usual \llama{} and \tower{} models were used in this scenario. For reference, \mosaic{} obtains an average AUC of 0.956 over all generators without adversarial attacks.}
    \label{tab:mosaic_robustness}
\end{table*}

\section{Complexity Improvements \label{ann:compute improvements}}
 Our algorithm currently processes each text in approximately 10 seconds on NVIDIA 32G V100 GPUs, and twice as fast on 80G A100 GPUs. Runtime optimization is an area that should be improved in future work. Below, we outline limitations of our system and propose potential improvements :
In \mosaic{}, the texts are processed one-by-one by the LLMs. Each model is loaded onto a separate GPU, and the logits are moved to a central device for performing operations such as Blahut-Arimoto, perplexity, and cross-entropy calculations, after which the final score is computed. This setup has several inefficiencies. For instance, transferring logits to a central device introduces a significant bottleneck. Additionally, while calculations are performed on one GPU, the remaining ones remain idle, resulting in suboptimal use of resources.

A more efficient method would involve computing the logits for all texts in parallel, storing them across different GPUs, and performing subsequent calculations concurrently. An even more streamlined solution would involve loading all models onto a single GPU using quantized or distilled versions, thus eliminating the need to transfer logits across devices.

While these optimizations are promising, they have not been implemented in this work, as we focus on the algorithmic methodology rather than runtime efficiency.

\section{A Systematic Study of all Potential Combinations}
Tables~\ref{tab:all-falcon-auc}, \ref{tab:all-falcon-tpr5}, \ref{tab:all-llama-auc} and \ref{tab:all-llama-tpr} display the whole study of all the potential combinations of the four models in our ensembles for every generator in the RAID dataset. 
\begin{table*}[htbp]
    \centering
    \resizebox{\textwidth}{!}{
    \begin{tabular}{|l|c|c|c|c|c|c|c|c|c|c|c|c|}
        \toprule
        Method & chatgpt & cohere-chat & cohere & gpt2 & gpt3 & gpt4 & llama-chat & mistral-chat & mistral & mpt-chat & mpt & \textbf{Average}\\
        \midrule
        Binoculars 0 1 & 0.99559 & 0.97838 & 0.97799 & 0.93460 & 0.99786 & 0.95868 & 0.99996 & 0.99919 & 0.91932 & 0.99733 & 0.95045 & 0.97358 \\ \hline
        FastDetectGPT 0 1 & 0.99487 & 0.96940 & 0.97366 & 0.94662 & 0.99375 & 0.96133 & 0.99922 & 0.99662 & 0.92746 & 0.99365 & 0.96846 & 0.97500\\ \hline
        Binoculars 0 2 & 0.94267 & 0.87411 & 0.85786 & 0.38375 & 0.97858 & 0.68632 & 0.98387 & 0.92952 & 0.59812 & 0.95399 & 0.64461 & 0.80304\\ \hline
        FastDetectGPT 0 2 & 0.86155 & 0.74420 & 0.77979 & 0.49724 & 0.82475 & 0.70021 & 0.94406 & 0.80315 & 0.64382 & 0.77470 & 0.75213 & 0.75687\\ \hline
        Binoculars 0 3 & 0.98818 & 0.92769 & 0.90545 & 0.41349 & 0.98959 & 0.85435 & 0.99703 & 0.98274 & 0.66670 & 0.98825 & 0.69790 & 0.85558\\ \hline
        FastDetectGPT 0 3 & 0.94627 & 0.82562 & 0.83573 & 0.52014 & 0.87708 & 0.82838 & 0.98374 & 0.91869 & 0.69212 & 0.89539 & 0.78809 & 0.82830\\ \hline
        Binoculars 1 0 & 0.70925 & 0.79631 & 0.73389 & 0.49105 & 0.93632 & 0.35652 & 0.88678 & 0.83358 & 0.66604 & 0.82795 & 0.71482 & 0.72296\\ \hline
        FastDetectGPT 1 0 & 0.61166 & 0.61442 & 0.64173 & 0.59703 & 0.68329 & 0.37331 & 0.82227 & 0.65597 & 0.68932 & 0.55185 & 0.80423 & 0.64046\\ \hline
        Binoculars 1 2 & 0.65047 & 0.67101 & 0.60557 & 0.20704 & 0.87062 & 0.29262 & 0.82634 & 0.66793 & 0.43608 & 0.69845 & 0.47293 & 0.58173\\ \hline
        FastDetectGPT 1 2 & 0.51206 & 0.43272 & 0.50416 & 0.36579 & 0.47719 & 0.34305 & 0.69285 & 0.43790 & 0.52221 & 0.35996 & 0.64190 & 0.48089\\ \hline
        Binoculars 1 3 & 0.85179 & 0.76976 & 0.66409 & 0.22159 & 0.90261 & 0.44492 & 0.93861 & 0.85577 & 0.46290 & 0.88676 & 0.49854 & 0.68158\\ \hline
        FastDetectGPT 1 3 & 0.64419 & 0.50397 & 0.52859 & 0.37228 & 0.49602 & 0.42459 & 0.76787 & 0.56583 & 0.52658 & 0.49533 & 0.64131 & 0.54241\\ \hline
        Binoculars 2 0 & 0.97183 & 0.97627 & 0.97972 & 0.82349 & 0.99761 & 0.89893 & 0.99819 & 0.99345 & 0.94065 & 0.99375 & 0.96866 & 0.95841\\ \hline
        FastDetectGPT 2 0 & 0.97335 & 0.96636 & 0.97593 & 0.85763 & 0.99305 & 0.91412 & 0.99851 & 0.98910 & 0.94400 & 0.98469 & 0.98023 & 0.96154\\ \hline
        Binoculars 2 1 & 0.99459 & 0.98569 & 0.98585 & 0.91758 & 0.99925 & 0.96204 & 0.99993 & 0.99942 & 0.95250 & 0.99732 & 0.97588 & \textbf{0.97910}\\ \hline
        FastDetectGPT 2 1 & 0.99470 & 0.97686 & 0.98179 & 0.93533 & 0.99594 & 0.96306 & 0.99993 & 0.99799 & 0.95402 & 0.99441 & 0.98457 & \textbf{0.97987}\\ \hline
        Binoculars 2 3 & 0.99417 & 0.98038 & 0.97641 & 0.68528 & 0.99709 & 0.96864 & 0.99807 & 0.99378 & 0.86997 & 0.99491 & 0.90954 & 0.94257\\ \hline
        FastDetectGPT 2 3 & 0.99618 & 0.97637 & 0.97700 & 0.72250 & 0.99601 & 0.97237 & 0.99997 & 0.99405 & 0.88457 & 0.99509 & 0.93609 & 0.95002\\ \hline
        Binoculars 3 0 & 0.85527 & 0.93737 & 0.95387 & 0.79709 & 0.99687 & 0.52522 & 0.99323 & 0.96276 & 0.89972 & 0.96142 & 0.94402 & 0.89335\\ \hline
        FastDetectGPT 3 0 & 0.86263 & 0.93726 & 0.95519 & 0.81322 & 0.99419 & 0.52588 & 0.99398 & 0.96045 & 0.91163 & 0.95339 & 0.95770 & 0.89687\\ \hline
        Binoculars 3 1 & 0.91470 & 0.92920 & 0.92923 & 0.88144 & 0.99245 & 0.60599 & 0.99881 & 0.98299 & 0.87891 & 0.97938 & 0.92556 & 0.91079\\ \hline
        FastDetectGPT 3 1 & 0.92361 & 0.93678 & 0.93578 & 0.89008 & 0.99040 & 0.62426 & 0.99792 & 0.98310 & 0.88995 & 0.97919 & 0.93025 & 0.91648\\ \hline
        Binoculars 3 2 & 0.84538 & 0.90754 & 0.91513 & 0.64079 & 0.99382 & 0.54105 & 0.98113 & 0.89614 & 0.76557 & 0.93707 & 0.83912 & 0.84207\\ \hline
        FastDetectGPT 3 2 & 0.85196 & 0.90944 & 0.91961 & 0.64439 & 0.99140 & 0.53714 & 0.98236 & 0.89891 & 0.78080 & 0.93265 & 0.85672 & 0.84594 \\
        \bottomrule
    \end{tabular}}
    \caption{AUROC on RAID for all configurations of Falcons for Binoculars and FastDetectGPT. Configurations are indicated by the index of the models used : \falcon-7b[0], \falcon-7b-instruct[1], \falcon-40b[2], \falcon-40b-instruct[3].}
    \label{tab:all-falcon-auc}
\end{table*}

\begin{table*}[htbp]
    \centering
    \resizebox{\textwidth}{!}{
    \begin{tabular}{|l|c|c|c|c|c|c|c|c|c|c|c|c|}
        \toprule
        Method & chatgpt & cohere-chat & cohere & gpt2 & gpt3 & gpt4 & llama-chat & mistral-chat & mistral & mpt-chat & mpt & \textbf{Average}\\
        \midrule
        Binoculars 0 1 & 0.98900 & 0.91600 & 0.90500 & 0.69000 & 0.99200 & 0.80400 & 1.00000 & 0.99600 & 0.62100 & 0.99400 & 0.72900 & 0.87600\\ \hline
        FastDetectGPT 0 1 & 0.99000 & 0.88800 & 0.89700 & 0.78700 & 0.98100 & 0.84400 & 1.00000 & 0.99000 & 0.71500 & 0.98400 & 0.85000 & 0.90236\\ \hline
        Binoculars 0 2 & 0.70200 & 0.55800 & 0.38200 & 0.00900 & 0.94900 & 0.08000 & 0.97600 & 0.62700 & 0.00500 & 0.80600 & 0.01300 & 0.46427\\ \hline
        FastDetectGPT 0 2 & 0.26200 & 0.12900 & 0.18500 & 0.00200 & 0.21400 & 0.03200 & 0.62800 & 0.10600 & 0.00800 & 0.10600 & 0.01900 & 0.15373\\ \hline
        Binoculars 0 3 & 0.97900 & 0.72900 & 0.57400 & 0.01000 & 0.98000 & 0.35000 & 0.99900 & 0.95900 & 0.02400 & 0.98000 & 0.03000 & 0.60127\\ \hline
        FastDetectGPT 0 3 & 0.72500 & 0.33100 & 0.29500 & 0.00200 & 0.32900 & 0.21000 & 0.96200 & 0.49900 & 0.01900 & 0.42200 & 0.04100 & 0.34864\\ \hline
        Binoculars 1 0 & 0.10100 & 0.31800 & 0.14200 & 0.01000 & 0.68700 & 0.00800 & 0.54300 & 0.21500 & 0.02000 & 0.28400 & 0.03200 & 0.21455\\ \hline
        FastDetectGPT 1 0 & 0.04700 & 0.07400 & 0.11400 & 0.01400 & 0.10500 & 0.00600 & 0.37900 & 0.05400 & 0.05300 & 0.04800 & 0.10200 & 0.09055\\ \hline
        Binoculars 1 2 & 0.03100 & 0.08000 & 0.03200 & 0.00600 & 0.34500 & 0.00500 & 0.35700 & 0.01700 & 0.00300 & 0.07900 & 0.00700 & 0.08745\\ \hline
        FastDetectGPT 1 2 & 0.00800 & 0.00100 & 0.01000 & 0.00000 & 0.01000 & 0.00400 & 0.00600 & 0.00000 & 0.00100 & 0.00200 & 0.00300 & 0.00409\\ \hline
        Binoculars 1 3 & 0.30600 & 0.29200 & 0.07200 & 0.00500 & 0.50200 & 0.01400 & 0.66700 & 0.23400 & 0.00300 & 0.46300 & 0.01000 & 0.23345\\ \hline
        FastDetectGPT 1 3 & 0.01600 & 0.00500 & 0.01200 & 0.00000 & 0.01400 & 0.00400 & 0.04100 & 0.00200 & 0.00100 & 0.00200 & 0.00400 & 0.00918\\ \hline
        Binoculars 2 0 & 0.89200 & 0.92800 & 0.94100 & 0.35000 & 0.99600 & 0.57700 & 0.99900 & 0.98500 & 0.74700 & 0.98600 & 0.85900 & 0.84182\\ \hline
        FastDetectGPT 2 0 & 0.89500 & 0.88200 & 0.92200 & 0.51400 & 0.97300 & 0.69400 & 0.99700 & 0.96100 & 0.78600 & 0.94400 & 0.91200 & 0.86182\\ \hline
        Binoculars 2 1 & 0.98700 & 0.95000 & 0.95300 & 0.58700 & 0.99700 & 0.85900 & 1.00000 & 0.99700 & 0.78400 & 0.99500 & 0.87700 & \textbf{0.90782}\\ \hline
        FastDetectGPT 2 1 & 0.98500 & 0.90900 & 0.92800 & 0.71900 & 0.98100 & 0.85300 & 1.00000 & 0.99200 & 0.79700 & 0.98300 & 0.91800 & \textbf{0.91500}\\ \hline
        Binoculars 2 3 & 0.99300 & 0.93100 & 0.92400 & 0.14300 & 0.99400 & 0.89700 & 1.00000 & 0.98800 & 0.52800 & 0.99500 & 0.60100 & 0.81764\\ \hline
        FastDetectGPT 2 3 & 0.99500 & 0.90800 & 0.91600 & 0.28600 & 0.98500 & 0.91300 & 1.00000 & 0.98500 & 0.63400 & 0.98700 & 0.76400 & 0.85209\\ \hline
        Binoculars 3 0 & 0.49300 & 0.82800 & 0.84400 & 0.39300 & 0.99200 & 0.05500 & 0.97800 & 0.86700 & 0.62400 & 0.85700 & 0.77400 & 0.70045\\ \hline
        FastDetectGPT 3 0 & 0.53400 & 0.78700 & 0.83700 & 0.48700 & 0.98000 & 0.09600 & 0.97500 & 0.83800 & 0.68500 & 0.79500 & 0.83700 & 0.71373\\ \hline
        Binoculars 3 1 & 0.71400 & 0.81700 & 0.80800 & 0.56500 & 0.96900 & 0.15100 & 0.99500 & 0.94600 & 0.59400 & 0.94000 & 0.73600 & 0.74864\\ \hline
        FastDetectGPT 3 1 & 0.70100 & 0.78800 & 0.77800 & 0.61100 & 0.95200 & 0.16400 & 0.99300 & 0.91900 & 0.62900 & 0.90700 & 0.77400 & 0.74691\\ \hline
        Binoculars 3 2 & 0.48200 & 0.73100 & 0.72300 & 0.13800 & 0.98900 & 0.04300 & 0.93800 & 0.66900 & 0.30000 & 0.77800 & 0.39800 & 0.56264\\ \hline
        FastDetectGPT 3 2 & 0.51000 & 0.70700 & 0.72900 & 0.21200 & 0.97200 & 0.09400 & 0.93500 & 0.65400 & 0.38900 & 0.71500 & 0.55800 & 0.58864\\ 
        \bottomrule
    \end{tabular}}
    \caption{TPR@5\%FPR on RAID for all configurations of Falcons for Binoculars and FastDetectGPT. Configurations are indicated by the index of the models used : \falcon-7b[0], \falcon-7b-instruct[1], \falcon-40b[2], \falcon-40b-instruct[3].}
    \label{tab:all-falcon-tpr5}
\end{table*}

\begin{table*}[htbp]
    \centering
    \resizebox{\textwidth}{!}{
    \begin{tabular}{|l|c|c|c|c|c|c|c|c|c|c|c|c|}
        \toprule
        Method & chatgpt & cohere-chat & cohere & gpt2 & gpt3 & gpt4 & llama-chat & mistral-chat & mistral & mpt-chat & mpt & \textbf{Average}\\
        \midrule
        Binoculars 0 1 & 0.98868 & 0.98521 & 0.97708 & 0.77221 & 0.99868 & 0.95950 & 1.00000 & 0.99325 & 0.87059 & 0.99826 & 0.89373 & \textbf{0.94884}\\ \hline
        FastDetectGPT 0 1 & 0.98843 & 0.98087 & 0.97859 & 0.80717 & 0.99594 & 0.96678 & 1.00000 & 0.98990 & 0.89063 & 0.99503 & 0.92696 & 0.95639\\ \hline
        Binoculars 0 2 & 0.78407 & 0.95574 & 0.96148 & 0.80034 & 0.99533 & 0.75383 & 0.98365 & 0.93690 & 0.89692 & 0.96825 & 0.93239 & 0.90626\\ \hline 
        FastDetectGPT 0 2 & 0.78205 & 0.95023 & 0.96258 & 0.82922 & 0.99061 & 0.76525 & 0.98419 & 0.93287 & 0.91102 & 0.95666 & 0.95267 & 0.91067\\ \hline 
        Binoculars 0 3 & 0.82575 & 0.90990 & 0.91668 & 0.50181 & 0.98496 & 0.62274 & 0.97900 & 0.88801 & 0.76507 & 0.94356 & 0.81524 & 0.83207\\ \hline 
        FastDetectGPT 0 3 & 0.77696 & 0.83897 & 0.88423 & 0.57974 & 0.93333 & 0.64056 & 0.95657 & 0.80964 & 0.79026 & 0.82991 & 0.87509 & 0.81048\\ \hline
        Binoculars 1 0 & 0.58200 & 0.79002 & 0.75954 & 0.70742 & 0.97505 & 0.49584 & 0.83473 & 0.69413 & 0.72506 & 0.78803 & 0.74389 & 0.73597\\ \hline
        FastDetectGPT 1 0 & 0.58753 & 0.80557 & 0.76156 & 0.66687 & 0.97223 & 0.45940 & 0.82525 & 0.71417 & 0.70701 & 0.81873 & 0.70547 & 0.72944\\ \hline
        Binoculars 1 2 & 0.56796 & 0.75465 & 0.76687 & 0.72228 & 0.96559 & 0.50900 & 0.80903 & 0.70925 & 0.72264 & 0.74014 & 0.75186 & 0.72902\\ \hline
        FastDetectGPT 1 2 & 0.54410 & 0.77559 & 0.77060 & 0.68639 & 0.96576 & 0.45812 & 0.78384 & 0.71097 & 0.70689 & 0.77309 & 0.71989 & 0.71775\\ \hline
        Binoculars 1 3 & 0.51082 & 0.68785 & 0.71171 & 0.51711 & 0.94521 & 0.37595 & 0.74094 & 0.55979 & 0.60934 & 0.66094 & 0.63689 & 0.63241\\ \hline
        FastDetectGPT 1 3 & 0.50535 & 0.67341 & 0.70523 & 0.50097 & 0.91377 & 0.36301 & 0.73958 & 0.55247 & 0.60627 & 0.64651 & 0.63589 & 0.62204\\ \hline
        Binoculars 2 0 & 0.95832 & 0.93661 & 0.92122 & 0.64083 & 0.99498 & 0.80933 & 0.99431 & 0.97987 & 0.78711 & 0.99023 & 0.78671 & 0.89087\\ \hline
        FastDetectGPT 2 0 & 0.95451 & 0.91550 & 0.92841 & 0.71961 & 0.98027 & 0.84700 & 0.99345 & 0.96768 & 0.84736 & 0.96719 & 0.87606 & 0.90882\\ \hline
        Binoculars 2 1 & 0.99466 & 0.97298 & 0.95712 & 0.66708 & 0.99810 & 0.95333 & 0.99997 & 0.99584 & 0.79541 & 0.99853 & 0.79537 & 0.92076\\ \hline
        FastDetectGPT 2 1 & 0.99318 & 0.95637 & 0.95732 & 0.73729 & 0.98816 & 0.96584 & 0.99996 & 0.99052 & 0.83830 & 0.99041 & 0.86718 & 0.93496\\ \hline
        Binoculars 2 3 & 0.93911 & 0.90875 & 0.87909 & 0.45877 & 0.98512 & 0.71763 & 0.98668 & 0.95586 & 0.69021 & 0.97678 & 0.69611 & 0.83583\\ \hline
        FastDetectGPT 2 3 & 0.90842 & 0.84184 & 0.85347 & 0.55726 & 0.91478 & 0.75533 & 0.97874 & 0.90606 & 0.75215 & 0.89647 & 0.81150 & 0.83418\\ \hline
        Binoculars 3 0 & 0.95895 & 0.95099 & 0.95582 & 0.73085 & 0.99658 & 0.83482 & 0.99527 & 0.98233 & 0.88173 & 0.98795 & 0.90806 & 0.92576\\ \hline
        FastDetectGPT 3 0 & 0.94743 & 0.92731 & 0.94888 & 0.78517 & 0.98916 & 0.86043 & 0.99348 & 0.96569 & 0.90673 & 0.96440 & 0.95181 & 0.93095\\ \hline
        Binoculars 3 1 & 0.99569 & 0.98342 & 0.97868 & 0.74037 & 0.99948 & 0.96954 & 0.99999 & 0.99752 & 0.87714 & 0.99773 & 0.89556 & 0.94865\\ \hline
        FastDetectGPT 3 1 & 0.99418 & 0.96904 & 0.97333 & 0.79378 & 0.99494 & 0.97381 & 0.99999 & 0.99281 & 0.90178 & 0.99130 & 0.94086 & \textbf{0.95689}\\ \hline
        Binoculars 3 2 & 0.93622 & 0.95978 & 0.95667 & 0.81153 & 0.99729 & 0.84474 & 0.99411 & 0.98110 & 0.91501 & 0.98647 & 0.94575 & 0.93897\\ \hline
        FastDetectGPT 3 2 & 0.93160 & 0.94640 & 0.95190 & 0.84095 & 0.99159 & 0.86386 & 0.99445 & 0.97235 & 0.92340 & 0.97124 & 0.96553 & 0.94121\\
        \bottomrule
    \end{tabular}}
    \caption{AUROC on RAID for all configurations of our \llama models for Binoculars and FastDetectGPT. Configurations are indicated by the index of the models used : \llama-2-7b[0], \llama-2-7b-chat[1], \towerbase-7B-v0.1[2], \towerbase-13B-v0.1[3].}
    \label{tab:all-llama-auc}
\end{table*}

\begin{table*}[htbp]
    \centering
    \resizebox{\textwidth}{!}{
    \begin{tabular}{|l|c|c|c|c|c|c|c|c|c|c|c|c|}
        \toprule
        Method & chatgpt & cohere-chat & cohere & gpt2 & gpt3 & gpt4 & llama-chat & mistral-chat & mistral & mpt-chat & mpt & \textbf{Average}\\
        \midrule
        Binoculars 0 1 & 0.95300 & 0.93700 & 0.89800 & 0.17300 & 0.99300 & 0.77200 & 1.00000 & 0.97200 & 0.38900 & 0.99500 & 0.42500 & \textbf{0.77336}\\\hline
        FastDetectGPT 0 1 & 0.94900 & 0.91600 & 0.90900 & 0.32400 & 0.98500 & 0.84300 & 1.00000 & 0.95800 & 0.55400 & 0.98300 & 0.64800 & \textbf{0.82445}\\ \hline 
        Binoculars 0 2 & 0.46000 & 0.85900 & 0.87500 & 0.31200 & 0.99000 & 0.28000 & 0.94300 & 0.78100 & 0.55700 & 0.88500 & 0.69000 & 0.69382\\ \hline 
        FastDetectGPT 0 2 & 0.44700 & 0.83100 & 0.87800 & 0.44800 & 0.97300 & 0.35300 & 0.94400 & 0.76400 & 0.65600 & 0.82000 & 0.80900 & 0.72027\\ \hline
        Binoculars 0 3 & 0.37100 & 0.66500 & 0.63700 & 0.01200 & 0.97500 & 0.13700 & 0.93500 & 0.51100 & 0.09400 & 0.73500 & 0.10800 & 0.47091\\ \hline
        FastDetectGPT 0 3 & 0.21900 & 0.40900 & 0.55900 & 0.04800 & 0.68700 & 0.11800 & 0.78400 & 0.26400 & 0.22700 & 0.27900 & 0.40200 & 0.36327\\ \hline
        Binoculars 1 0 & 0.07400 & 0.38000 & 0.29600 & 0.17300 & 0.90300 & 0.01600 & 0.22600 & 0.17500 & 0.19600 & 0.27100 & 0.20600 & 0.26509\\ \hline
        FastDetectGPT 1 0 & 0.05800 & 0.34500 & 0.27900 & 0.15200 & 0.88500 & 0.01500 & 0.21600 & 0.14600 & 0.18600 & 0.22800 & 0.19200 & 0.24564\\ \hline
        Binoculars 1 2 & 0.06600 & 0.30200 & 0.29200 & 0.17400 & 0.84700 & 0.02400 & 0.15200 & 0.16900 & 0.19000 & 0.18100 & 0.21500 & 0.23745 \\ \hline
        FastDetectGPT 1 2 & 0.06400 & 0.28700 & 0.28800 & 0.16500 & 0.83700 & 0.02100 & 0.16000 & 0.16000 & 0.20000 & 0.16300 & 0.21900 & 0.23309\\ \hline
        Binoculars 1 3 & 0.01300 & 0.11900 & 0.13500 & 0.01300 & 0.71500 & 0.00700 & 0.05500 & 0.02000 & 0.02600 & 0.05400 & 0.03300 & 0.10818\\ \hline
        FastDetectGPT 1 3 & 0.01200 & 0.08000 & 0.12300 & 0.01800 & 0.49900 & 0.00600 & 0.05900 & 0.01300 & 0.03500 & 0.01900 & 0.06800 & 0.08473\\ \hline 
        Binoculars 2 0 & 0.76200 & 0.71000 & 0.54000 & 0.03100 & 0.99000 & 0.23200 & 0.99000 & 0.89700 & 0.08200 & 0.96300 & 0.06200 & 0.56900\\ \hline
        FastDetectGPT 2 0 & 0.74800 & 0.66300 & 0.66400 & 0.11700 & 0.89900 & 0.36900 & 0.98200 & 0.84400 & 0.28500 & 0.83200 & 0.36000 & 0.61482\\ \hline
        Binoculars 2 1 & 0.98900 & 0.85400 & 0.76300 & 0.03200 & 0.99300 & 0.71100 & 1.00000 & 0.98600 & 0.12800 & 0.99700 & 0.08600 & 0.68536\\ \hline
        FastDetectGPT 2 1 & 0.97100 & 0.79600 & 0.77500 & 0.11400 & 0.93900 & 0.81400 & 1.00000 & 0.96400 & 0.25500 & 0.96300 & 0.33900 & 0.72091\\ \hline
        Binoculars 2 3 & 0.69700 & 0.63100 & 0.45000 & 0.01400 & 0.97500 & 0.17000 & 0.98600 & 0.76700 & 0.04000 & 0.91400 & 0.02300 & 0.51518\\ \hline
        FastDetectGPT 2 3 & 0.50400 & 0.43000 & 0.43400 & 0.01500 & 0.58300 & 0.17100 & 0.94100 & 0.50800 & 0.11900 & 0.48200 & 0.16600 & 0.39573\\ \hline 
        Binoculars 3 0 & 0.80100 & 0.81800 & 0.81400 & 0.11700 & 0.99700 & 0.32500 & 0.99300 & 0.92500 & 0.35600 & 0.97700 & 0.44000 & 0.68755\\ \hline
        FastDetectGPT 3 0 & 0.75500 & 0.74300 & 0.80400 & 0.28300 & 0.96600 & 0.45100 & 0.97900 & 0.85300 & 0.57000 & 0.85400 & 0.76600 & 0.72945\\ \hline
        Binoculars 3 1 & 0.98800 & 0.92500 & 0.90100 & 0.11000 & 0.99900 & 0.83900 & 1.00000 & 0.99300 & 0.32000 & 0.99400 & 0.37100 & 0.76727\\ \hline
        FastDetectGPT 3 1 & 0.97700 & 0.85900 & 0.88400 & 0.27200 & 0.98100 & 0.88200 & 1.00000 & 0.96900 & 0.54000 & 0.97200 & 0.70000 & 0.82145\\ \hline
        Binoculars 3 2 & 0.70000 & 0.85700 & 0.86600 & 0.26900 & 0.99400 & 0.35600 & 0.98100 & 0.91200 & 0.54500 & 0.96400 & 0.69700 & 0.74009\\ \hline
        FastDetectGPT 3 2 & 0.69200 & 0.79100 & 0.83400 & 0.41200 & 0.96900 & 0.45800 & 0.97400 & 0.86700 & 0.64900 & 0.85300 & 0.81300 & 0.75564\\ 
        \bottomrule
    \end{tabular}}
    \caption{TPR@5\%FPR on RAID for all configurations of our \llama models for Binoculars and FastDetectGPT. Configurations are indicated by the index of the models used : \llama-2-7b[0], \llama-2-7b-chat[1], \towerbase-7B-v0.1[2], \towerbase-13B-v0.1[3].}
    \label{tab:all-llama-tpr}
\end{table*}

\end{document}